\documentclass[a4paper]{article}[12pts]  
\usepackage[english]{babel}
\usepackage[utf8x]{inputenc}
\usepackage[T1]{fontenc}
\usepackage{array,booktabs}% http://ctan.org/pkg/{array,booktabs}
\normalsize
\usepackage{graphicx}

\usepackage{natbib}

\usepackage[a4paper,top=1in,bottom=1in,left=1in,right=1in,marginparwidth=1in]{geometry}
\usepackage{natbib}

\usepackage{setspace}
\usepackage{amssymb}
\usepackage{amsthm}
\usepackage{amsmath}
\DeclareMathOperator*{\argmax}{arg\,max}
\DeclareMathOperator*{\argmin}{arg\,min}
\usepackage{amsfonts, mathtools}
\usepackage{algorithm}
\usepackage{algorithmic}
\usepackage{bm}
 \usepackage{bbm}
\usepackage{comment}
\usepackage{graphicx}
\usepackage{multirow, booktabs}
\usepackage{caption}
\usepackage{subcaption}
\usepackage[colorinlistoftodos]{todonotes}
\usepackage[english]{babel}
\usepackage[utf8x]{inputenc}
\usepackage[T1]{fontenc}

\newcommand{\E}{\mathbb{E}}

\newcommand{\prob}{\mathbb{P}}

\newcommand{\qedwhite}{\hfill \ensuremath{\Box}}
\newtheorem{remark}{Remark}
\newtheorem{theorem}{Theorem}

\newtheorem{corollary}{Corollary}
\newtheorem{lemma}{Lemma}

\newtheorem{assumption}{Assumption}

 \def\bibfont{\small}%
 \bibpunct[, ]{[}{]}{,}{n}{}{,}%
\newcommand{\hz}[1]{{\color{blue} [HZ: {#1}]}}

\usepackage[colorlinks=true,breaklinks=true,bookmarks=true,urlcolor=blue,
     citecolor=blue,linkcolor=blue,bookmarksopen=false,draft=false]{hyperref}

\title{Learning to Sell a Focal-ancillary Combination}

\author{Hanzhao Wang \and Xiaocheng Li \and Kalyan Talluri}
\date{\small
Imperial College Business School, Imperial College London\\
$\{$h.wang19, xiaocheng.li, kalyan.talluri$\}$@imperial.ac.uk\\
}

\begin{document}

\maketitle

\onehalfspacing
\begin{abstract}%
A number of products are sold in the following sequence:  First a focal product is shown, and if the customer purchases, one or more ancillary products are displayed for purchase.
A prominent example is the sale of an airline ticket, where first the flight is shown, and when chosen, a number of ancillaries such as cabin or hold bag options, seat selection,
insurance etc. are presented.   The firm has to decide on a sale format---whether to sell them in sequence unbundled, or together as a bundle---and how to price the focal and ancillary products, separately or as a bundle.
Since the ancillary is considered by the customer only after the purchase of the focal product, the sale strategy chosen by the firm creates an information and learning dependency between the products: for instance,  offering only a bundle would preclude learning customers' valuation for the focal and ancillary products individually.
In this paper we study  learning strategies for such focal and ancillary item combinations under the following scenarios: (a) pure unbundling to all customers, (b) personalized mechanism, where, depending on some observed
features of the customers, the two products are presented and priced as a bundle or in sequence, (c) initially unbundling (for all customers), and switch to bundling (if more profitable) permanently once during the horizon.
We design pricing and decisions algorithms for all three scenarios, with regret upper bounded by $O(d  \sqrt{T} \log T)$, and an optimal switching time for the third scenario.
\end{abstract}%
%\KEYWORDS{deterministic inventory theory; infinite linear programming duality;
%  existence of optimal policies; semi-Markov decision process; cyclic schedule}
%\MSCCLASS{Primary: 90B05; secondary: 90C40, 90C90}
%\ORMSCLASS{Primary: Inventory/production: deterministic multi-item;
%  secondary: dynamic programming/optimal control: deterministic
%  semi-Markov; programming: infinite dimensional}
%\HISTORY{Received November 20, 2003; revised March 8, 2004, and March 26, 2004.}

%\MSCCLASS{}
%\ORMSCLASS{Primary: ; secondary: }

\section{Introduction}
A number of products are sold in conjunction with one or more ancillary products.  The most prominent example is an airline flight where once you purchase the ticket, you are given options on choosing a seat, buying
luggage, insurance etc. (ancillaries).  A hotel booking can lead to additional charges such as parking fees, high-speed Wifi etc.
In e-commerce applications, warranties, packaging or gift-wrapping, refund and return policies, credit offers, and shipping and card-payment options can also be considered ancillaries. Indeed ancillary revenue is considered as an important source of additional revenue in many industries.

There are multiple ways firms price ancillaries in relation to the focal product.  They can price the ancillary significantly higher than cost (for instance airline seat choice or luggage options)
or significantly less than cost (free warranties or return policy), often called the ``hold-up" strategy or the ``give-away" strategy.   
Thus studying sale mechanisms for
such a focal-ancillary combination is of great interest.  In this paper we consider the learning and pricing aspects of the problem for such focal-ancillary product combinations.

It should be clear that the sale of a focal-ancillary combination has some idiosyncracies---there is a certain order and dependence in how they are sold.
The firm can choose to sell them either bundled together or as a sequential process with two distinct prices where the consumer is offered the ancillary only after purchasing the focal product.   If the firm has good
estimates of the parameters of the demand model it may be able to make a decision on bundled or unbundled offers up-front, but if it has to learn the parameters then the choice of the sale
mechanism would influence the learning of the parameters.  For instance, if the firm were to follow a strict bundling strategy from the beginning, it may learn the valuations for the bundle as a whole, but will be unable to determine if an unbundling strategy
is better as it cannot learn the
valuations for the focal and ancillary products separately.  Similarly, even if it were to follow an unbundling strategy, the price of the focal item would influence the number of samples
it can collect, and hence the rate for learning the ancillary product valuations.

In this paper we study learning strategies for such focal and ancillary item combinations under the following scenarios: (a) pure unbundling to all customers, (b) personalized mechanism where depending on some observed
covariates of the customers, the two products are presented as a bundle or in sequence, (c) the firm can learn by unbundling, but has to make a decision to fix bundling for all customers.
We design pricing and decisions algorithms with regret upper bounded by $O(d\sqrt{T} \log T)$, and an optimal switching time for the third scenario .

The contributions of the paper are as follows:
\begin{enumerate}
\item For the pure unbundling case, we give an algorithm that sets prices for the focal and ancillary separately while learning the parameters of the demand model.  With no distributional assumptions on the
co-variates of the arriving customers we show the regret of the algorithm is
$O\left(d \sqrt{T} \log T\right)$.  Under an i.i.d assumption on the distribution of the co-variates, this regret bound can be improved to $O\left(d^2\log^2(T)/q^*\right)$, albeit with a dependence on a quantity $q^*$ representing the expected purchase probability of the focal product under
optimal pricing with the true parameters. For above two regret bounds, we present two different ways to bound the single-period regret: ``ex-ante'' and ``ex-post''. The former is the bound before observing the focal product purchase or non-purchase while the later is after observing it or knowing the focal product's purchase probability. Further, to the best of our knowledge, this is the first paper introducing the learning rate $q^*$ in the regret bound for learning and pricing problems.
\item When the mechanism can be personalized to each customer depending on their observed covariates, offering either a bundle or the focal-ancillary sequence, we give an algorithm that, again without making any
distributional assumptions on the covariates, achieves an $O\left( d  \sqrt{T}  \log T \right)$ regret. When designing the algorithm, we borrow the idea of \textit{Information Directed Sampling(IDS)} (\citet{russo2018learning}) for balancing the learning and earning when deciding the selling strategy for exploration. To the best of our knowledge, this is the first time using IDS in learning and revenue management problem.
\item Finally, when the firm wants to fix a single mechanism for all customers, but does not know the parameters of the demand model and wants to determine when to bundle, we give an
algorithm that learns and
switches to bundling with a regret of $ O\left(d\log T \sqrt{\frac{T}{q^*}}\right)$ under an i.i.d assumption on the distribution of the covariates.
\end{enumerate}

The rest of the paper is organized as follows: In \S\ref{sec:litRev} we survey the relevant literature, in \S\ref{sec:model} we set out the model and the notations, and in the three subsequent sections analyze our learning algorithms
for the three different scenarios---the pure unbundling case in \S\ref{sec_pure_unbundle}, the personalized mechanism in \S\ref{sec_general} and the optimal switching to bundling in \S\ref{sec_one_switch}.

\section{Literature Review}\label{sec:litRev}
In this section we survey the literature relevant to our paper.
\subsection{Pricing of focal-ancillary products}
The economic motivation for bundling and its use for price discrimination of a heterogenous set of consumers is well studied (\citet{stigler1963united, adams1976commodity, McAfee89}).
Further light on the effectiveness of bundling is thrown by recent theoretical works (\citet{hart2017approximate, li2013revenue}) and
empirical research (\citet{chu2011bundle}).

As we mentioned earlier, sale of focal and ancillary products is distinct because of the asymmetric dependence of the latter on the former.
Add-on pricing and the economics of lock-in, hold-up and giveaways have been examined in the Industrial Organization   literature from various angles (\citet{Shapiro95,Ellison05, Fang06}).
The analysis of \citet{Gomes18} shows the monopolist's tendency to extract rents from the ancillary product is tempered by the probability of missed sales.

In the Revenue Management literature, the concern is on optimal pricing of either the bundled products or the pricing sequence for selling the focal and ancillary (\citet{Cui18, Wang19, Allon11,Brueckner15}).
As far as we know we are the first paper to consider learning for this type of sale of products.  The dependency in sales and information creates some new technical issues that we tackle.

\subsection{Bandits and sequential recommendations}
Our techniques and regret framework comes from the learning literature and we collect some relevant references here.
In the cascading bandits model, the agent offers $K$ items to
the user,  the user examines the list from top to bottom and chooses the first attractive item, a learning variant of the cascade model where the agent needs to learn the $K$ most attractive items.
This bandit setting was first introduced by \citet{kveton2015cascading} and then extended
in \citet{li2016contextual, kveton2015combinatorial}.

There are important differences between our model and the  cascading bandits model: (i) In cascading bandits, the user examines a product only
when they reject all previous ones, but for us, the ancillary item is considered only when they buy the focal item; (ii)
The reward function in cascading bandits is the probability that the customer will accept at least one product, but in our model the objective is to maximize revenue; (iii)
The decision variable in cascading bandits is the set of recommended items and their positions, while the decision variables for us are the prices as well as whether to bundle, and if so when (in the third variant that we analyze).

\citet{chen2021revenue} consider the revenue maximization problem with unknown parameters and a cascading model for customer behavior.
However, the decision variables in this work are still rankings of
offered items, and  the customer is assumed to purchase at most one item.   Another work, \citet{song2018dynamic}, is based on a similar customer
purchase process model as ours and also focuses on optimal dynamic pricing.
However, their focus is on a situation with finite inventory, full information and no customer
covariates; thus the dynamic pricing is driven by the state of inventory, while the dynamic pricing in our
work is to trade-off learning the demand and earning by choosing the best selling strategy and optimal pricing.

We would also like to mention the recent literature on feature-based dynamic pricing (\citet{cohen2020feature, javanmard2019dynamic, ban2021personalized}) which
consider a single product. As in these papers, our demand model for the focal and
ancillary items are based on customer features, but in addition to   a more complicated sale mechanism, we also consider the case where the arriving customer features are not necessarily i.i.d.

\section{Model}\label{sec:model}
A seller has two products to sell, a focal product and an ancillary product, over a horizon of $T$ time periods.
The focal product is the main revenue driver, while the ancillary product is a closely related product that may bring the costumer additional utility and further boost the
total earnings of the seller.
The distinguishing feature of this combination is that the ancillary product can be purchased/used only after the focal product
has been purchased. The two products are presented in that order to the customer.
From a revenue management perspective, there are two decisions to make (for the seller):
(i) whether to \textit{bundle} the two items, and (ii) how to \textit{price}.
If they are bundled, the seller will post one single price for the bundle, and the customer will decide to purchase either the bundle (both of the products) or nothing.
If they are sold unbundled, the two products have their own individual prices, and the customer can choose to buy only the focal product, and if they purchase the focal, decide on purchasing the ancillary product.

We denote the chosen selling strategy in each period $t=1,...,T$ as $A_t\in\{b,u\}$,
where $b$ denotes bundling and $u$ denotes unbundling. 
We represent the prices under unbundling (subscript $u$) and bundling (subscript $b$) as $(\left[p_f,p_a\right]_u,p_b)$,
where $p_f, p_a$ are the focal and ancillary product prices.
The prices may vary over different time periods, indexed by a subscript $t$.

We assume $p_f,p_a,p_b\in [\underline{p},\bar{p}]$, although our analysis can be easily generalized to the case when the feasible sets are different for the three prices.
%When $A_t=u$, it means that the bundling price $p_b$ is not effective; similarly, when $A_t=b,$ it means the unbundling prices $p_f$ and $p_a$ are not effective.
We assume both products have zero marginal costs so that the revenue equals to the profit and there are no capacity constraints.

Customer context is a $d$-dimensional feature vector $x_t\in \mathcal{X}\subseteq \mathbb{R}^d$ that describes the
selling environment and the customer features at time period $t$, where $\mathcal{X}$ is the set
of possible contexts.  We assume customers do not form rational expectations and are not strategic on their time of arrival.
In contrast to the literature on feature-based dynamic pricing (\citet{ javanmard2019dynamic, ban2021personalized}),
we also consider the case without assumptions on the statistical structure on the $x_t$'s.

Denote the demand vector of customer at a certain period as $\left[d_f,d_a\right]_u, d_{b}$, where $d_{f},d_{a} , d_{b} \in \{0,1\}$, with 1 indicating a sale and 0, no sale.
When the seller chooses unbundling
$d_{b}=0$ and in addition if $d_f=0$, then $d_a=0$.  Likewise, $d_{f}=d_{a}=0$ when the seller chooses bundling.

The demand model for a single time period for an observation of $x$ and prices $[p_{f},p_{a}]_u,p_{b}$ is as follows:
When the selling strategy is unbundling, the utility of the arriving customer for the focal product is
$x^\top \theta_f^* + \epsilon_f$
and that for the ancillary product $x^\top \theta_a^* + \epsilon_a$
where the random variables $\epsilon_f$ and $\epsilon_a$ capture utility unexplained by the feature vector $x$.

For focal and ancillary
products, there are two fixed unknown response vectors $\theta_f^*,\theta_a^*$ from a bounded set $\Theta \subset \mathbb{R}^d$ that we assume does not change over time.
Thus the demand in the unbundled setting in a single period is
\begin{equation*}
[d_{f},d_{a}]_u=\begin{cases}
 [1,1], &\text{if }x^\top \theta_f^* + \epsilon_f\geq p_{f} \text{ and }  x^\top \theta_a^* + \epsilon_a\geq p_{a},\\
 [1,0], &\text{if }x^\top \theta_f^* + \epsilon_f\geq p_{f} \text{ and }  x^\top \theta_a^* + \epsilon_a <  p_{a},\\
[0,0], &\text{if } x^\top \theta_f^* + \epsilon_f < p_{f},
\end{cases}
\end{equation*}
(and the bundling demand $d_b=0$ given the selling strategy is pure unbundling).
%So the customer first compares their utility of the focal product against its price $p_f$ and decides whether to purchase the focal product.
%Conditional on a focal item purchase, the customer will then compare their utility of the ancillary product
%against the price $p_a$ and decide whether to purchase the ancillary item also.
This model of consumer decision-making is standard in the economics literature (see for instance \S5 of \citet{Gomes18}) and assumes the firm does not reveal the
ancillary price before the focal item purchase (economists refer to this as ``drip" pricing) and  the consumer
does not form rational expectations of the ancillary price; so the consideration set is not affected by the combined price.

Alternatively, if the selling strategy is bundling, the utility of the arriving customer for the bundle is modeled by
$x^\top \theta_b^* + \epsilon_b$
where $\theta_b^* \coloneqq \theta^*_f+\theta^*_a$ denotes the response vector for the product bundle, and the randomness in utility
$\epsilon_b\coloneqq \epsilon_f+\epsilon_a$. %It does not hurt to assume $\theta_b^*\in \Theta$ as well.

In other words, we are making the following assumption:
\begin{assumption}[Utility Additivity]
We assume that the utility of the bundle is equal to the sum of the utilities of the focal and ancillary product.
\label{assp_add}
\end{assumption}
On the other hand, the results in this paper continue to hold if $\epsilon_b$ takes a different distribution than $\epsilon_f+\epsilon_a$.

The demand under the bundling sale mechanism hence is
\begin{equation*}
d_{b}=\begin{cases}
 1, &\text{if } x^\top \theta_b^* + \epsilon_{b}\geq p_{b},  \\
0, &\text{otherwise}
\end{cases}
\end{equation*}
with $d_f=d_a=0$ (as the selling strategy is bundling).

Let $F_{\epsilon_f}$ and $F_{\epsilon_a}$ denote the cumulative distribution function of $\epsilon_f$ and $\epsilon_a$. Let $F_{\epsilon_b}$ denote the cumulative distribution function of $\epsilon_b$ (which as we mentioned needs not to be the convolution of $F_{\epsilon_f}$ and $F_{\epsilon_a}$). Then the probability model under the two mechanisms for the realized demand with feature $x$ is as follows for the unbundling case:
    \begin{equation*}
    [d_{f},d_{a}]_u=\begin{cases}
 [1,1], &\text{w.p. } \left(1-F_{\epsilon_f}(p_{f}-  x^\top \theta_f^*)\right)\left(1-F_{\epsilon_a}(p_{a}-x^\top \theta_a^*)\right),\\
 [1,0], &\text{w.p. } \left(1-F_{\epsilon_f}(p_{f}-x^\top \theta_f^*)\right)F_{\epsilon_a}(p_{a}-x^\top \theta_a^*),\\
[0,0], &\text{w.p. } F_{\epsilon_f}(p_{f}-x^\top \theta_f^*).
\end{cases}
\end{equation*}
and for the bundling case:
    \begin{equation*}
    d_{b}=\begin{cases}
 1, &\text{w.p. }1-F_{\epsilon_b}(p_{b}-x^\top \theta_b^* ),  \\
0, &\text{w.p. } F_{\epsilon_b}(p_{b}-x^\top \theta_b^*).
\end{cases}
\end{equation*}

We now turn to the multi-period setting.
First, we consider the random utility shock as idiosyncratic noises, thus independent over time.
For simplicity, we also assume they are identically distributed according to the law of $F_{\epsilon_f}, F_{\epsilon_a}$ and $F_{\epsilon_b}$.
We use $(\epsilon_{t,f}, \epsilon_{t,a},\epsilon_{t,b})$ to denote the utility shock at time $t$.
Accordingly, we use $[p_{t,f}, p_{t,a}]_u, p_{t,b}$ and $[d_{t,f},d_{t,a}]_u,d_{t,b}$ to denote the price and realized demand at time $t$.
In this way, all the above model formulations can be cast into a
multi-period form by replacing the price, demand, feature, and utility shock with their time-varying counterparts.
From a technical perspective, the prices and the selling strategy (to bundle or not) are decisions to make and accordingly, the seller collects observations of customer demands. Through these observations, the seller aims to learn the unknown fixed response vector $(\theta_f^*, \theta_a^*)$ so as to better understand the demand mechanism and hence to optimize the prices and the selling strategy.

\subsection{Assumptions}
We summarize our boundedness structure on the parameter $\theta$ ($=\theta_f,\theta_a \text{ or }\theta_b $), the feature vector $x$, the utility shocks and the prices.
\begin{assumption}[Boundedness] We assume
\begin{itemize}
    \item[(a)] There exists a known $\bar{\theta}>0$ and $\Theta=\{\theta\in \mathbb{R}^d: \|\theta\|_2\leq \bar{\theta}\}$.
    \item[(b)] For any covariate vector $x\in \mathcal{X}$,  $\|x\|_2\leq 1$.
    \item[(c)] For focal, ancillary, and bundle product, the seller is always allowed with a price range $[\underline{p},\bar{p}]$ under all possible $x$ and $\theta$. Furthermore, for $\forall \theta\in\Theta, x\in \mathcal{X}$, the optimal prices to maximize expected revenue fall in the range $[\underline{p},\bar{p}]$.
\end{itemize}
\label{assp_bound}
\end{assumption}

Next, the following assumption describes the structure we impose on the random
utility shocks $\epsilon_f, \epsilon_a$ and $\epsilon_b$. The assumption seems restrictive but in essence it only
requires some dispersion in the randomness. This type of assumption, together with the distribution knowledge,
is often treated as part of the customer choice model and has been standard in the feature-based dynamic pricing literature
(\citet{javanmard2017perishability, javanmard2019dynamic, ban2021personalized}).
The emphasis of this line of literature as in our paper is on the interplay of parameter learning and pricing decision-making.
Thus the assumption on the underlying distribution aims to ensure a moderate speed of learning.

\begin{assumption}[Distribution] We assume the random utility shock distributions $ F_{\epsilon_f} ,F_{\epsilon_a},F_{\epsilon_b}$ are continuous functions with density function $f_{\epsilon_f},f_{\epsilon_a},f_{\epsilon_b}$. Moreover, we assume that the $F_{(\cdot)}(v)$'s are strictly increasing in the interval $\left[\underline{p}-\bar{\theta}, \Bar{p}+\bar{\theta}\right]$, and that both $F_{(\cdot)}(v)$ and $1-F_{(\cdot)}(v)$ are log-concave in $v$, and there exist constants $B$ and $B'$ such that $$B=\max_{v\in \left[\underline{p}-\bar{\theta}, \Bar{p}+\bar{\theta}\right]}\max\left\{f_{\epsilon_f}(v),f_{\epsilon_a}(v),f_{\epsilon_b}(v)\right\},$$ and $$B'=\max_{v\in \left[\underline{p}-\bar{\theta}, \Bar{p}+\bar{\theta}\right]}\max\left\{|f_{\epsilon_f}'(v)|,|f_{\epsilon_a}'(v)|,|f_{\epsilon_b}'(v)|\right\}.$$
\label{assp_distr}
\end{assumption}
Our assumption adopts that of \citet{javanmard2019dynamic}, and we note that log-concavity is satisfied by common probability distributions like normal, uniform and logistic.

Based on the distribution functions, the following two quantities will be critical in determining the learning rate of the problem:
\begin{align*}
   \nu&=\inf_{|v|\leq \bar{p}+\bar{\theta}}\left\{\min_{i\in \{f,a,b\} }\{ -\log''F_{\epsilon_i}(v),-\log''(1-F_{\epsilon_i}(v))\}\right\}, \\
   \mu&=\sup_{|v|\leq \bar{p}+\bar{\theta}}\left\{\max_{i\in \{f,a,b\}}\{ -\log'F_{\epsilon_i}(v),-\log'(1-F_{\epsilon_i}(v))\}\right\}.
\end{align*}
Under Assumption~\ref{assp_bound} and~\ref{assp_distr}, we know that $\nu$ and $\mu$ are well-defined. As we will see in the later sections, they govern the strong convexity and smoothness of the log-likelihood function.

As stated earlier, the main results of our paper do not rely on any statistical assumptions on the feature vectors $x_t$'s.
At certain parts of our paper, we will also show that if the $x_t$'s are sampled i.i.d. from some distribution, the algorithm
performance can be improved. For that purpose, we introduce a distributional assumption on $x_t$, even if we do not
require it for our main discussion. We will use capital letter $X$ for random variables, and small one for realized values.

\begin{assumption}
\label{assp_X}
Feature vectors $X_t$'s are generated independently from a fixed unknown distribution $F_X$ with a bounded support  $\mathcal{X}$ in $\mathbb{R}^d$.
We denote by $\Sigma=\mathbb{E}\left[XX^\top\right]$ the second moment matrix of distribution $F_X$,
and we assume that $\Sigma$ is a positive definite matrix, and there exist
$\bar{\lambda}, \underline{\lambda}$ such that $\lambda_{\max}\left(X_tX_t^\top \right)\leq \bar{\lambda}$ almost surely  for all $t$ and $\lambda_{\min}\left(\Sigma\right)\geq \underline{\lambda}$.
\end{assumption}

For a  positive definite matrix $\Sigma$, define the $\Sigma$-norm of a vector $x$ as,
$$\|x\|_\Sigma \coloneqq \sqrt{x^\top \Sigma x}.$$

\subsection{Selling strategy with known parameters} \label{sec_revenue_func}
In this section we collect  results on the optimal pricing and expected demand for a single period when the true parameters are known. For brevity, we denote the deterministic part of the product valuation (given a feature $x$)
as $v_f=  x^\top  \theta_f$, $v_a= x^\top  \theta_a$, and $v_b= x^\top \theta_b = v_f+v_a$.

 %Let $p= (p_f,p_a,p_b)$ be a feasible  price vector and denote

The seller's expected single-period revenue under the bundling strategy is given by
\begin{equation*}
    r_b(p_b; v_b) \coloneqq p_b\left(1-F_{\epsilon_b}(p_b-v_b)\right).
\end{equation*}
and under the unbundling strategy,
\begin{equation*}
    r_u(p_f,p_a; v_f,v_a) \coloneqq \left(p_f+p_a(1-F_{\epsilon_a}(p_a-v_a))\right)\left(1-F_{\epsilon_f}(p_f-v_f)\right).
\end{equation*}

For a cumulative distribution function $F_{\epsilon_i}(v)$, $ i \in \{f,a,b\}$, define  %CANT SEE THE DOT; KTT
$$
g_{i}(v)\coloneqq v+\phi_{i}^{-1}(-v),
$$
where $\phi_{i}(v)=v-(1-F_{\epsilon_i}(v))/f_{\epsilon_i}(v)$ is the virtual valuation function.
%Here the symbol $\cdot$ in the subscripts stands for the focal product $f$, the ancillary product $a$, or the bundle $b$.
The revenue-maximizing pricing rule can be defined using the functions $g_i(\cdot)$, $ i \in \{f,a,b\}$  as follows.
\begin{lemma}\label{lem:static1}[Optimal pricing for single item \citet{javanmard2019dynamic}]
The revenue-maximizing price is given by
$$p_b^*(v_b)=g_{b}(v_b)$$
for the bundling product and
$$p_a^*(v_a)=g_{a}(v_a)$$
for the ancillary product.
\end{lemma}
The focal product's optimal price however has to be derived differently as it
depends on the expected revenue of the ancillary product $r_a$.
%Intuitively, the expected revenue of the ancillary product can be viewed as part of the deterministic evaluation of the focal product.
In the following, we write the optimal price for the focal product as a function of $v_f$ and $r_a$.

\begin{lemma}\label{lem:static2}[Optimal pricing for focal item]
The revenue-maximizing price for the focal product when $r_a$ is the expected revenue of the ancillary product, is given by
$$p_f^*(v_f,r_a)=g_f(v_f+r_a)-r_a.$$
\end{lemma}

\proof{}
\noindent
Given $v_f$ and $r_a$, we can write the expected revenue for unbundling as below:
$$(p_f +r_a)(1-F_{\epsilon_f}(p_f-v_f)).$$

Thus, by the first order condition, the optimal $p_f^*(v_f,r_a)$ should satisfy
$$p_f^*=\frac{1-F_{\epsilon_f}(p_f^*-v_f)}{f_{\epsilon_f}(p_f^*-v_f)}-r_a.$$
Rearranging it we get
$$\phi_f(p^*_f-v_f)+v_f+r_a=0.$$
Thus the optimal pricing
$$p_f^*(v_f,r_a)=g_f(v_f+r_a)-r_a.$$
 \qedwhite

%\subsubsection{Optimal revenue and selling strategy}
When the seller prices the products according to the optimal functions in Lemmas~\ref{lem:static1} and \ref{lem:static2}, we can define the optimal expected revenue  as a function of the underlying valuation $v$, as
$r_i(v)\coloneqq g_i(v)\left(1-F_{\epsilon_i}(g_i(v)-v)\right), i \in \{f,a,b\}.$
%As earlier, the subscript $\cdot$ may stand for focal product $f$, ancillary product $a$, or the bundle $b$.

Thus the optimal revenue function under the bundling strategy is
\begin{equation*}
    r^*_b(v_b)=r_b(v_b)
\end{equation*}
while the optimal revenue function for the unbundling strategy is
\begin{equation*}
    r^*_u(v_f,v_a)=r_f(v_f+r_a(v_a)).
\end{equation*}
The optimal selling strategy, for the single-period case, with known parameters, hence is simply given as $b$ if $r^*_b(v_b) > r^*_u(v_f,v_a)$ and $u$ otherwise.

\subsection{Performance measure}
In the following sections, we will study three settings:
(i) \textit{Pure unbundling} refers to the setting where only the unbundling strategy is allowed and the seller only needs to decide the prices of the focal and ancillary products. (ii) \textit{Personalized mechanism} where both the selling mechanism and pricing are up to the seller to decide, and both can be personalized based on the feature vector of the customer. (iii) \textit{One-switch} when the selling strategy is initially unbundling, and same for all customers and the seller may switch to bundling at some point for the rest of the horizon.

Under all three settings, we assume the parameters $\theta^*_f, \theta^*_a, \theta^*_b$ are unknown and measure the performance of an online policy $\pi$ by the notion of \textit{regret}:
$$\text{Regret}^{\pi} \coloneqq r^* - \E[R^{\pi}]$$
where $r^*$ refers to the optimal expected revenue obtained when we know the parameters and can optimize both the selling strategy when available as well as the price accordingly for each arriving customer. $R^{\pi}$ is the revenue from the policy $\pi$.  We remark that the expectation is over the utility shocks $\epsilon_f, \epsilon_a$ and $\epsilon_b$ and potential randomness in policy $\pi$. We leave the definition of regret in an open form here and will specify it later under each context of the three settings. 

Both the revenue functions are defined for an arbitrary fixed sequence of the co-variate vectors $x_t$.
In case Assumption~\ref{assp_X} holds, where the customer co-variate vectors are assumed to be generated i.i.d from a known distributions, then the expectation would be over the co-variate distribution also.

We develop algorithms and derive regret upper bounds to evaluate the algorithm performance. In presenting the upper bound, we will treat parameters such as $B, B', \nu, \mu, \bar{\theta}$ (pertaining to the known utility shock distributions) as constants and emphasize the dependence of regret on the parameters like  $d$, $T$ and other related quantities.

\begin{comment}
\subsection{Notations}

\begin{enumerate}
    \item $\mathcal{T}_{t, f/a/b}$: the subset of  $\{1,..,t-1\}$ that contains all periods gathering focal/ancillary/bundle item samples. By gathering samples, we mean at that period we get the feedback if the customer purchase the corresponding item ($d_{t,f/a/b}\neq \emptyset$).
    \item $\lambda_{\min}(M)$: the minimal eigenvalue of a matrix $M$.
    \item $\theta^*_f,\theta^*_a,\theta^*_b$ true parameters. Given $x_t$, we also use $V_{t,f}:=x_t^\top \theta^*_f$, $V_{t,a}:=x_t^\top \theta^*_a$, $V_{t,b}:=x_t^\top \theta^*_b$.
\end{enumerate}

\xiaocheng{The following statement should be postponed, we have touched the statistical assumptions:
\begin{itemize}
    \item Further, if $F$ and $1-F$ are both log-concave, $p_f(v_f,r_a)$ is strictly decreasing in $r_a$ and strictly increasing in $v_f$.
    We note that the monotonicity coincides with our intuitions: the larger expected revenue from ancillary will decrease the focal item's price in order to increase the opportunity to consider ancillary item; the larger value of focal item will however increase its price for earning more from its value.
\end{itemize}}
\end{comment}

\section{Pure Unbundling} \label{sec_pure_unbundle}
We first study learning under a \textit{pure unbundling} setting where only the unbundling strategy is available.
%The setting is depicted as the diagram below. Though it does not
%involve the choice of selling strategy, it may still be of
%independent interest for practical applications.
Our discussion of this setting illustrates some technical and operational
differences between two-product sequential pricing and single-product pricing.
Also, it prepares us for the more complicated setting where both the bundling and unbundling strategies are available.

Mathematically, the setting fixes the strategy to be unbundling $A_t=u$ for all $t=1,...,T$.
Prices are personalized, so the seller observes $x_t$ and sets the prices of both the focal and ancillary products, $p_{t,f}$ and $p_{t,a}$.

%\vspace{0.1in}
%
%\begin{center}
%\begin{tikzpicture}[
%rdsquarednode/.style={rounded corners, draw=black!100, fill=red!0, thick, minimum size=5mm, node distance=2cm, text width=3.3cm,  text centered},
%]
%%Nodes
%\node[rdsquarednode]      (1)                              {Observe $x_t$};
%\node[rdsquarednode]        (2)       [right=of 1] {Decide $p_{t,f}$ and $p_{t,a}$};
%
%%Lines
%\draw[->] (1.east) -- (2.west) ;
%\end{tikzpicture}
%\end{center}
%
%\vspace{0.1in}

The regret of a certain policy/algorithm $\pi$ is
\begin{equation}
    \text{Reg}^{\pi}_T(\bm{X}) \coloneqq  \sum_{t=1}^T r^*_u(x_{t}^\top \theta_f^*,x_{t}^\top \theta_a^*)- \mathbb{E}\left[\sum_{t=1}^T r_u(P_{t,f},P_{t,a};x_{t}^\top \theta_f^*,x_{t}^\top \theta_a^*)\right]
\end{equation}
%\sum_{t=1}^T r^*_u(x_{t}^\top \theta_f^*,x_{t}^\top \theta_a^*)-\E_{\epsilon} \left[\sum_{t=1}^T R_u(p_{t,f},p_{t,a};x_{t}^\top \theta_f^*,x_{t}^\top \theta_a^*)\right]=
where $\bm{X} = \{x_1,...,x_T\}$ encapsulates the covariates of all time periods, and the expectation is taken with respect to the potential random pricing $P_{t,f},P_{t,a}$ induced by the policy $\pi$.
%Here $V_{t,f}=x_{t}^\top \theta_f^*$ and $V_{t,a}=x_{t}^\top \theta_a^*$ denote the (conditionally) deterministic part of the customer valuation.
The functions $r^*_u$ and $r_u$ are defined in \S\ref{sec_revenue_func}. Specifically, $r^*_u$ denotes the optimally achievable expected revenue given the knowledge of the true parameters $\theta_f^*$ and $\theta_a^*$, and thus it represents the clairvoyant optimal revenue. The function $r_u$ denotes the expected revenue given the prices and the customer valuations.
%We note that there is no expectation taken for the first summation in the regret definition as the utility shock randomness is already absorbed in the function $r^*_u$ and $r_u$, and the feature set $\bm{X}$ bears no statistical structure in general.

\subsection{Pricing Algorithm}

We first give some intuition into our pricing algorithm. Recall that the focal purchase and the ancillary purchase happen in a sequential manner. In other words, if
the customer at certain time $t$ chooses not to purchase the focal product, the seller would not be able to make an observation on the customer's decision for the ancillary product.
Technically, it means that the purchase history of focal product determines the observations/samples collection for the ancillary product.
This is a key difference from both bundle pricing as well as single-product dynamic learning and pricing problems.
The sequential structure means the seller should not over-price the focal product too often as it affects the sample collection rate (equivalently, the learning rate) of the ancillary demand model.
In addition, the samples used for estimating the ancillary demand model consist of only a subset of all the samples that may result in a degeneracy in the covariance structure.

Algorithm~\ref{alg_pure_b} describes our lower confidence bound (LCB)-based pricing policy for the unbundling setting.   At each time $t$, the algorithm estimates the parameters by the regularized maximum likelihood estimation (MLE).

We introduce the log-likelihood function in a form generally applicable to any setting or product. Specifically, let data set $\mathcal{D}=\{(p_t,x_t, d_{t}), t\in\mathcal{T}\}$ denote a set of available observations with price $p_t$ (for either focal, ancillary, or bundle products), customer feature vector $x_t$, utility shock distribution $F_{\epsilon}$ ($\epsilon=\epsilon_f,\epsilon_a \text{ or } \epsilon_b$ depending on the given data set), and realized demand $d_t$. Let $\mathcal{T}$ denote the index set of the observations.

Denote the data sets $\mathcal{D}_{t,f} = \{(p_{t',f}, x_{t'},d_{t',f}), t'=1,...,t-1\}$ and
$\mathcal{D}_{t,a} = \{(p_{t',a}, x_{t'},d_{t',a}), t'\in\mathcal{T}_{t,a}\}$ where the time index
set $$\mathcal{T}_{t,a} \coloneqq \{t'|d_{t',f}=1,t'=1,...,t-1\}$$ denotes the time periods when a focal purchase takes place.

The log-likelihood function is
$$LL(\theta;\mathcal{D}) \coloneqq \sum_{t\in \mathcal{T},d_{t}=1}\log(1-F_\epsilon(p_{t}-x_{t}^\top\theta ))+\sum_{t\in \mathcal{T},d_{t}=0}\log(F_\epsilon(p_{t}-x_{t}^\top\theta)).$$
For time $t\ge 2$, the regularized MLEs for the focal product and for the ancillary product are given by
\begin{equation*}
    \hat{\theta}_{t,f}\coloneqq \argmin_{\theta\in \Theta} \;\; -LL(\theta;\mathcal{D}_{t,f})+\lambda\nu \|\theta\|_2^2,
\end{equation*}
\begin{equation*}
    \hat{\theta}_{t,a}\coloneqq\argmin_{\theta\in \Theta}\;\; -LL(\theta;\mathcal{D}_{t,a})+\lambda\nu \|\theta\|_2^2,
\end{equation*}
where $\lambda$ is the regularization parameter to be specified and $\nu$ is defined earlier following Assumption~\ref{assp_distr}.

Next, we construct a ``data-driven'' confidence set following the standard treatment in the linear bandits literature \citet{lattimore2020bandit}.
For a regularization parameter $\lambda >0$, define the following design matrices for each period
$$\Sigma_{t,f}\coloneqq\sum_{t'=1}^t x_{t'}x_{t'}^\top+\lambda I, \ \ \Sigma_{t,a}\coloneqq\sum_{t'\in \mathcal{T}_{t,a}}x_{t'}x_{t'}^\top+\lambda I,$$
where $I$ is a $d$-dimensional identity matrix.

Define the confidence sets
$$\Theta_{t,f}\coloneqq\left\{\theta\in \Theta: \left\|\hat{\theta}_{t,f}-\theta\right\|_{\Sigma_{t-1,f}}\leq \beta(\Sigma_{t-1,f})\right\}, \ \ \Theta_{t,a}\coloneqq\left\{\theta\in \Theta: \left\|\hat{\theta}_{t,a}-\theta\right\|_{\Sigma_{t-1,a}}\leq \beta(\Sigma_{t-1,a})\right\},$$
where the function $\beta$   is defined by
\begin{equation*}
\label{beta_eq}
  \beta(\Sigma)\coloneqq 2\sqrt{\lambda}\bar{\theta}+ \frac{2\mu}{\nu}\sqrt{2\log T+\log\left(\frac{\det \Sigma}{\lambda^d}\right)}
\end{equation*}
with $\bar{\theta}$ as given in Assumptions~\ref{assp_bound} and $\mu, \nu$ following Assumptions~\ref{assp_distr}.
Discussions on the property of the estimators and justifications on the confidence set construction are referred to in Appendix~\ref{sec_MLE_analysis}.

The algorithm uses the confidence sets to construct LCB and UCB for the customer valuations of
the focal and ancillary product. The seller sets the prices based on the optimal pricing functions (in \S\ref{sec_revenue_func}) as if the
corresponding valuations are true. Specifically, the ancillary product's price comes directly from the estimator $\hat{\theta}_{t,a}$ and the focal
product's price adopts a lower confidence principle. The rationale is that when the seller has some uncertainty about the underlying model, a price lower than the
optimal price will lead to a sufficient number of focal purchases for learning the ancillary product's parameter at a good rate.
To this end, we point out that despite its name, the lower confidence pricing does not aim for \textit{exploration} and its sole purpose is to ensure the sample collection rate for the ancillary product.

\begin{algorithm}[ht!]
\caption{LCB Pricing for Pure Unbundling}
\label{alg_pure_b}
\begin{algorithmic}
\STATE{\textbf{Input:} Regularization parameter $\lambda$.}
\FOR{$t=1,...,T$}
\STATE{Compute the estimators $\hat{\theta}_{t,f}$ and $\hat{\theta}_{t,a}$ and their confidence sets $\Theta_{t,f}$ and $\Theta_{t,a}$}
\STATE{Observe feature $x_t$ and compute the UCB and LCB of the customer valuation:
$$\underline{v}_{t,f}=\min_{\theta\in\Theta_{t,f}} x_t^\top \theta,\quad \bar{v}_{t,a}=\max_{\theta\in\Theta_{t,a}} x_t^\top \theta, $$
}
\STATE{Set the price by
\begin{equation}
\label{LCB_pricing}
    p_{t,f}=p^*_f\left(\underline{v}_{t,f},r_a^*(\bar{v}_{t,a})\right)
\end{equation}
\begin{equation}
\label{LCB_pricing_2}
   p_{t,a}=p_a^*(x_t^\top \hat{\theta}_{t,a})
\end{equation}
where the optimal pricing functions $p^*_f$ and $p^*_a$ are given in \S\ref{sec_revenue_func}.
}
\ENDFOR
\end{algorithmic}
\end{algorithm}

We remark that the regularized MLE has two advantages over the standard MLE: first, it ensures well-definedness
of the optimization problem when there are not enough observations, say for the first few periods. Second, for the parameter estimation of the ancillary product, the regularization and the construction of the confidence set based on the sample design matrix $\Sigma_{t,\cdot}$ both help to overcome the ``selection bias'' induced by the focal purchase. Specifically, we recall that the samples used for estimating the ancillary parameters are those when the focal purchase takes place. It may happen that a subset of the customers (with certain covariate structure) tend to purchase the focal product more often than others which will twist the sample space used for estimating the ancillary parameters. Again, the intuition here is aligned with the literature on linear bandits (\citet{lattimore2020bandit}) where the selection bias is induced by the arm plays.

Let $$\eta=B+\bar{p}B'$$
and
$$\bar{\beta}=2\bar{\theta}+ \frac{2\mu}{\nu}\sqrt{2\log\left(T\right)+d\log\left(\frac{d +T}{d}\right)}.$$
\begin{theorem}
\label{UB_purepricing_2}
Under Assumptions~\ref{assp_add},~\ref{assp_bound},~\ref{assp_distr} and with the regularization parameter $\lambda = 1$, the regret of the Algorithm~\ref{alg_pure_b} for the pure
unbundling setting is bounded by $$2\bar{p}+6\sqrt{2}\bar{\beta}\sqrt{dT\log\left(\frac{d+T}{d}\right)}+ 2d\eta\bar{\beta}^2 \log\left(\frac{d+T}{d}\right) = O\left(d\sqrt{T}\log T\right).$$
\end{theorem}

\begin{theorem}
\label{UB_purepricing_1}
Under Assumptions~\ref{assp_add},~\ref{assp_bound},~\ref{assp_distr}, and \ref{assp_X}, and with the regularization parameter $\lambda = 1$, the regret of the Algorithm~\ref{alg_pure_b} for the pure unbundling setting is bounded by
$$2\bar{p}+\frac{288d\eta \bar{\beta}^2}{q^*}\log\left(\frac{d+T+1}{d}\right) = O\left(\frac{d^2\log^2T}{q^*}\right).$$
Here $q^*$ represents the expected purchasing probability of the focal product under an optimal pricing policy that knows the true parameters, i.e.,
$$q^*:=\mathbb{E}_X \left[1-F_{\epsilon_f}\left(p^*_f(X^\top\theta^*_f,r^*_a(X^\top\theta^*_a))-X^\top\theta^*_f\right)\right].$$
%where the expectation is taken with respect to the random covariates $X$.
\end{theorem}

The above two theorems provide two regret upper bounds for Algorithm~\ref{alg_pure_b} for pure unbundling.
Theorem~\ref{UB_purepricing_2} makes no assumptions on the covariates $x_t$'s while Theorem~\ref{UB_purepricing_1} assumes $X_t$'s are i.i.d. (Assumption~\ref{assp_X}).
In this sense, we can view Theorem~\ref{UB_purepricing_2} as a worst-case bound and Theorem~\ref{UB_purepricing_1} as a problem-dependent bound: the latter bears a dependence on the underlying distribution through the parameter $q^*$. Notably, the i.i.d. assumption reduces the regret's dependence on $T$ from $\sqrt{T}$ to $\log T$, but involves an extra parameter $q^*$. The parameter $q^*$ represents the rate/probability of focal purchase under the optimal pricing policy which knows the true parameters. When the true parameters are unknown, $q^*$ governs the learning rate of the ancillary product in our pricing policy.
%As we will see in \S\ref{sec_lower_bound}, the dependence on $1/q^*$ cannot be further improved. Intuitively, this is because if one desires for a larger learning rate for the ancillary product,
%it will inevitably hurt the focal's revenue.  %SECTION REMOVED? KTT Yes,removed HZ
In the following, we will elaborate a few key steps in deriving the regret bounds and postpone all proofs to Appendix~\ref{sec_proof_pure_unb}.

\begin{comment}
\paragraph{Remark:}
\begin{itemize}
    \item [(1)] None self-selection bias of $X_t$ implies no forced exploration: Since the estimators for $\theta_f$ and $\theta_a$ are based on i.i.d $X_t$, there has no self-selection bias and thus no need to apply forced explorations like in Algorithm 1.
    \item [(2)] Learning rate $q^*$ for ancillary item: $q^*$ is the expected purchasing probability of focal item at optimal pricing under true parameter, which is indeed the learning rate of ancillary item even we assume the ancillary item's information is unknown but the optimal price of focal item is known, which will be shown to match the lower regret bound.
    \item [(3)] LCB pricing for focal item: by applying LCB of optimal pricing at (\ref{LCB_pricing}), the algorithm can guarantee the expected learning rate of ancillary item larger than $q^*$ without causing too much regret.
\end{itemize}
\end{comment}

\subsection{Regret Analysis}

First, we define ``good'' events
$$\mathcal{E}_{f}=\left\{\theta_{f}^*\in \Theta_{t,f} \text{ for } t=1,...,T\right\},$$
$$\mathcal{E}_{a}=\left\{\theta_{a}^*\in \Theta_{t,a} \text{ for } t=1,...,T \right\},$$
under which the confidence sets cover the corresponding true coefficients throughout the entire horizon. we obtain the following probability bounds by applying a union bound to the result in Corollary~\ref{MLEbound}.
\begin{lemma}
We have
$$\prob(\mathcal{E}_{f} \cap \mathcal{E}_{a}) \ge 1 -2/T.$$
\label{lemma_high_prob}
\end{lemma}

Conditional on the good event $\mathcal{E}_{f} \cap \mathcal{E}_{a}$, the focal price $p_{t,f}$ used in Algorithm~\ref{alg_pure_b} will be a lower bound of the clairvoyant optimal price, regardless of the underlying $x_t$. In this way, the LCB price will encourage more focal purchases (equivalently, ancillary observations) than the optimal policy in a pathwise manner.

\begin{lemma}[Lower Bound Pricing]
\label{price_1}
Conditional on the event $\mathcal{E}_f\cap \mathcal{E}_a$, for $\forall t=1,...,T$,
$$p_{t,f}\leq p^*_{f}(x_t^\top\theta^*_{f},r^*_a(x_t^\top\theta^*_a))\quad$$
holds with probability $1$.
\end{lemma}

\begin{comment}
Define the parameter estimators for the corresponding LCB/UCB estimated valuation,
$$\underline{\theta}_{t,f}\coloneqq \argmin_{\theta\in\Theta_{t,f}} x_t^\top \theta,\quad \bar{\theta}_{t,a}\coloneqq \argmax_{\theta\in\Theta_{t,a}} x_t^\top \theta.$$
\end{comment}

\subsection{Two ways to upper bound single-period regret} Now we present two different upper bounds for the singe-period regret in Lemma~\ref{purepricing_singlereg_1} and Lemma~\ref{purepricing_singlereg_2}. Essentially, the single-period regret compares the expected revenue obtained by Algorithm~\ref{alg_pure_b} against the optimal revenue at time $t$,
$$\text{Reg}_t \coloneqq  r_u^*\left(x_t^\top \theta_f^*,x_t^\top \theta_a^*\right)-r_u\left(p_{t,f},p_{t,a};x_t^\top \theta_f^*,x_t^\top \theta_a^*\right).$$
We name the two upper bounds ``ex-ante" and ``ex-post" in that they take different perspectives in analysis---the first before
observing the focal purchase/non-purchase and the second after observing it.

Specifically,
note that the quantity $\|x_t\|_{\Sigma_{t-1,a}^{-1}}$ can be viewed as a proxy of our estimation error for both product valuations (the focal's estimation will always be more accurate than the ancillary's). Lemma~\ref{purepricing_singlereg_1} upper bounds the single-period regret by the quadratic of the error in valuation estimation.
This is a more conventional bound and it is aligned with existing analysis (\citet{broder2012dynamic, javanmard2019dynamic, ban2021personalized}) where
the revenue loss grows quadratically with the estimation error.

In comparison, Lemma~\ref{purepricing_singlereg_2} takes into account the binary demand structure of the problem where
$$q_{t,f}=1-F_{\epsilon_f}\left({p}_{t,f}-x_t^\top \theta_f^* \right)$$
represents the
focal product's purchase probability given covariate $x_t$. Through a coupling argument of the optimal policy and our policy (Algorithm~\ref{alg_pure_b}), the regret will only be incurred if there is a focal purchase. Conditional on the ex-post observation of a focal purchase, the regret contains a linear term which explains the effect of ancillary estimation error on the focal's revenue, and a quadratic term which accounts for the effect of the error on the ancillary's revenue. Given that $\|x_t\|_{\Sigma_{t-1,a}^{-1}}$ is very small (See Appendix~\ref{sec_MLE_analysis}), the linear term in Lemma~\ref{purepricing_singlereg_2} will be larger than the quadratic term in Lemma~\ref{purepricing_singlereg_1}, the effect of which could be offset by the extra term $q_{t,f}$ . It may be hard to argue which bound is tighter;  these two bounds will lead to different regret bounds in Theorem~\ref{UB_purepricing_2} and Theorem~\ref{UB_purepricing_1}.

\begin{lemma}[Ex-ante Single-Period Regret Bound]
\label{purepricing_singlereg_1}
Under $\mathcal{E}_f \cap \mathcal{E}_a$, we have the following bound for the single-period regret
$$\mathrm{Reg}_t \leq 144\eta \bar{\beta}^2\cdot \|x_t\|^2_{\Sigma_{t-1,a}^{-1}},$$
where $\eta=B+\bar{p}B'$ and $\bar{\beta}=2\sqrt{\lambda}\bar{\theta}+ \frac{2\mu}{\nu}\sqrt{2\log\left(T\right)+d\log\left(\frac{d\lambda +T}{d\lambda}\right)}$. The prices $p_{t,f}$ and $p_{t,a}$ are given by Algorithm~\ref{alg_pure_b}.
% and $V_{t,f}$ and $V_{t,a}$ are the true (deterministic) valuations at time $t$.
\end{lemma}

\begin{lemma}[Ex-post Single-Period Regret Bound]
\label{purepricing_singlereg_2}
Under $\mathcal{E}_f \cap \mathcal{E}_a$, we have the following bound for the single-period regret
$$\mathrm{Reg}_t \leq q_{t,f}\cdot\left(6\bar{\beta}\|x_t\|_{\Sigma^{-1}_{t-1,a}}+\eta\bar{\beta}^2\|x_t\|_{\Sigma^{-1}_{t-1,a}}^2\right),$$
where $\eta=B+\bar{p}B'$ and $\bar{\beta}=2\sqrt{\lambda}\bar{\theta}+ \frac{2\mu}{\nu}\sqrt{2\log\left(T\right)+d\log\left(\frac{d\lambda +T}{d\lambda}\right)}$. Here the prices $p_{t,f}$ and $p_{t,a}$ are given by Algorithm~\ref{alg_pure_b}.
%, $V_{t,f}$ and $V_{t,a}$ are the true (deterministic) valuations at time $t$
%and $q_{t,f}=1-F_{\epsilon_f}\left({p}_{t,f}-V_{t,f}\right)$ denotes purchase probability of the focal product.
\end{lemma}

The proofs of Theorem~\ref{UB_purepricing_2} and Theorem~\ref{UB_purepricing_1} build on the above two lemmas. Specifically, the lemmas represent the regret with an interplay between $x_t$ and $\Sigma_{t,a}$'s. Intuitively, if certain $x_t$ causes a large single-period regret, it will be an effective observation in terms of reducing $\Sigma_{t,a}$. The Elliptical Potential Lemma (Lemma~\ref{EPL_lemma}, see its origin from \citet{lai1982least}) precisely characterizes such a tradeoff. We refer to the detailed proofs for the results in this section to Appendix~\ref{sec_proof_pure_unb} and the proof of the two regret bound theorems to Appendix~\ref{sec_UB_purepricing}.

\begin{comment}
\subsection{Proof sketch}
\begin{itemize}
\item Smoothness of revenues: At time $t$, given $H_{t-1}$, by using 'LCB' pricing $p^*_f(\underline{v}_{t,f},\bar{v}_{t,a})$ and $p^*_a(\bar{v}_{t,a})$, the single period regret can be bounded by
$$O\left(\left|\underline{v}_{t,f}-V_{t,f}\right|^2+\left|\bar{v}_{t,a}-V_{t,a}\right|^2\right),$$
which can be further bounded under good events $\mathcal{E}_{f}\cap \mathcal{E}_a$ by
$$O\left(\bar{\beta}\left\|x_t\right\|^2_{V_{t-1,a}^{-1}} \right),$$
where $\bar{\beta}=O\left(d\log(T)\right)$ is a uniform upper bound of $\beta(V)$.
\item Efficient exploration of ancillary item: Notice that $X_{t'}$ are independent with $V_{t,a}$ for all $t'> t$ and are i.i.d samples. Under good events $\mathcal{E}_f\cap \mathcal{E}_a$,  by using LCB pricing and  the expected time interval of updating $V_{t,a}$, namely, the time interval of gathering one more ancillary sample, is bounded by $1/q^*$, where $q^*$ is the expected purchasing probability of focal item at optimal pricing under the true parameter. Further,  the total expected regret happened in this time interval can be bounded by $$O\left(\frac{\bar{\beta}}{q^*}\E\left[\|X_t\|^2_{V_{t-1,a}^{-1}}\right]\right).$$
\item Total expected regret: With  elliptical potential lemma, by summing $$\sum_{t=1}^T \E\left[\left\|X_t\right\|^2_{V_{t-1,a}^{-1}}\right]=\frac{\bar{\beta}}{q^*}\E\left[\sum_{t\in S_{a,T}} \left\|X_t\right\|^2_{V_{t-1,a}^{-1}}\right],$$
the total regret can be bounded by
$$O\left(\frac{d^2\log^2T}{q^*}\right).$$
 \end{itemize}
\end{comment}

\section{Personalized Mechanism}\label{sec_general}

Now we consider the main formulation where the seller has the option of choosing the selling strategy in each time period. Specifically, the selling strategy can be dependent on the covariate $x_t$. As shown in the diagram below, at each time $t$, the seller first observes the covariate $x_t$ and then decides the selling strategy and the price(s).

\bigskip

\begin{tikzpicture}[
rdsquarednode/.style={rounded corners, draw=black!100, fill=red!0,  thick, minimum size=5mm, node distance=2cm, text width=3cm, align=left, text centered},
]
%Nodes
\node[rdsquarednode]      (1)                  {Observe $x_t$};
\node[rdsquarednode]        (2)       [right=of 1] {Bundle or not?};

\node[rdsquarednode]        (3)       [right=of 2] {Decide price(s)};

%Lines
\draw[->] (1.east) -- (2.west) ;
\draw[->] (2.east) -- (3.west) ;
\end{tikzpicture}

\bigskip

Under this setting, the regret of a policy/algorithm $\pi$ can be defined by
\begin{equation*}
\text{Reg}_{T}^\pi(\bm{X})\coloneqq \sum_{t=1}^T \max\left\{r^*_u\left(x_t^\top \theta_f^*,x_t^\top \theta_a^*\right), r^*_b\left(x_t^\top \theta_b^*\right)\right\} -  \mathbb{E}\left[\sum_{t=1}^T R^{\pi}_t\right]
\end{equation*}
where $\bm{X}$ encapsulates all the covariates as before and the expectation is taken
with respect to utility shocks and the potential randomness in policy $\pi$.  $R^{\pi}_t$ which is the random revenue obtained by $\pi$ at time $t$.
%Here the expectation is taken with respect to $r_t$ which is the revenue obtained by $\pi$ at time $t$.
%For the first summation, $V_{t,f}, V_{t,a}$ and $V_{t,b}$ are the (conditionally) deterministic evaluation of the customer, and
The functions $r_u^*$ and $r_b^*$ are the optimal revenue functions defined in \S\ref{sec_revenue_func}. We compare the online cumulative revenue against a benchmark which knows the underlying parameters and picks the more profitable one among the two selling strategies.

\subsection{Algorithm}

To describe the algorithm, we first introduce some notation. Recall that $A_t$ denotes the selling strategy at time $t$. With little overload of notations, denote the sets of time periods
\begin{align*}
\mathcal{T}_{t,f} & = \{t'|A_{t'}=u,t'=1,...,t-1\},\\
\mathcal{T}_{t,a} & = \{t'|A_{t'}=u,d_{t',f}=1,t'=1,...,t-1\},\\
\mathcal{T}_{t,b} & = \{t'|A_{t'}=b,t'=1,...,t-1\}.
\end{align*}
Specifically, the time periods that the strategy of bundling or unbundling is applied contribute to the corresponding parameter estimation. As before, the effective samples for the estimation of the ancillary product require a realized purchase of the focal item. Then we can construct the same regularized MLE estimators as before by
\begin{align*}
\hat{\theta}_{t,f}&:= \argmin_{\theta\in \Theta} \;\; -LL(\theta;\{(p_{t',f},x_t,d_{t',f}),t'\in \mathcal{T}_{t,f}\})+\lambda\nu \|\theta\|_2^2,\\
\hat{\theta}_{t,a}&:=\argmin_{\theta\in \Theta} \;\; -LL(\theta;\{(p_{t',a},x_t,d_{t',a}),t'\in \mathcal{T}_{t,a}\})+\lambda\nu \|\theta\|_2^2,\\
\hat{\theta}_{t,b}&:=\argmin_{\theta\in \Theta}\;\; -LL(\theta;\{(p_{t',b},x_t,d_{t',b}),t'\in \mathcal{T}_{t,b}\})+\lambda\nu \|\theta\|_2^2,
\end{align*}
where $\lambda$ is the regularization parameter. The design matrices are defined by
$$\Sigma_{t,f}=\sum_{t'\in \mathcal{T}_{t,f}} x_{t'}x_{t'}^\top+\lambda I, \ \ \Sigma_{t,a}=\sum_{t'\in \mathcal{T}_{t,a}}x_{t'}x_{t'}^\top+\lambda I,\  \ \Sigma_{t,b}=\sum_{t'\in \mathcal{T}_{t,b}}x_{t'}x_{t'}^\top+\lambda I,$$
then we have the following confidence sets for the estimators
$$\Theta_{t,f}:=\left\{\theta\in \Theta: \left\|\hat{\theta}_{t,f}-\theta\right\|_{\Sigma_{t-1,f}}\leq \beta(\Sigma_{t-1,f})\right\},$$
$$\Theta_{t,a}:=\left\{\theta\in \Theta: \left\|\hat{\theta}_{t,a}-\theta\right\|_{\Sigma_{t-1,a}}\leq \beta(\Sigma_{t-1,a})\right\},$$
$$\Theta_{t,b}:=\left\{\theta\in \Theta: \left\|\hat{\theta}_{t,b}-\theta\right\|_{\Sigma_{t-1,b}}\leq \beta(\Sigma_{t-1,b})\right\},$$
where the function $\beta$ is defined by
$\beta(\Sigma)=2\sqrt{\lambda}\bar{\theta}+ \frac{2\mu}{\nu}\sqrt{2\log T+\log\left(\frac{\det \Sigma}{\lambda^d}\right)}
$
with $\bar{\theta}$, $\mu$ and $\nu$ defined in Assumptions~\ref{assp_bound} and~\ref{assp_distr}.

When both selling strategies are allowed, an alternative way to estimate the coefficient vector of the ancillary product is
\begin{equation}
\label{theta_a_estimate}
    \hat{\theta}'_{t,a}:=\hat{\theta}_{t,b}-\hat{\theta}_{t,f}.
\end{equation}
%Essentially, the result arises from the additivity of product utility in Assumption~\ref{assp_add}.

Algorithm~\ref{Alg2} extends the idea of Algorithm~\ref{alg_pure_b} to incorporate the option in choosing the selling strategy. Specifically, at each time $t$, it consists of two parts: the first part decides the selling strategy while the second part sets the price. When deciding the selling strategy, it compares the LCB revenue of each strategy with the UCB of the other. If one LCB is larger than the other UCB, it means for this feature $x_t,$ the seller can be quite confident about the superiority of one strategy and will adopt that strategy accordingly. Otherwise, the selling strategy will be chosen by \eqref{IDS_exploration}. Recall that $ \|x_t\|_{\Sigma^{-1}_{t-1,f}}$ and $\|x_t\|_{\Sigma^{-1}_{t-1,b}}$ represent the confidence level of the revenue of the focal and bundle product. The choice by \eqref{IDS_exploration} favors the strategy with larger uncertainty, and thus implements the ``exploration'' of the selling strategy. For the pricing part, if the seller chooses the unbundling strategy, then the problem reduces to the pure unbundling case of previous section and the pricing policy in Algorithm~\ref{alg_pure_b} can be applied. Alternatively, if the seller chooses the bundling strategy, it can be viewed as a single-product dynamic pricing problem and the certainty-equivalent pricing policy can be applied.

\begin{algorithm}[ht!]
\caption{Confidence-based Pricing Algorithm}
\label{Alg2}
\begin{algorithmic}
\STATE{\textbf{Input}: Regularization parameter: $\lambda$.}
\WHILE{$t=1,...,T$}
\STATE{\textcolor{blue}{\%\%Choose the strategy:}}
\STATE{After observing $x_t$, compute UCBs and LCBs for optimal revenues under both strategies by
$$ \bar{r}_{t,u}^*:=r_u^*\left(\bar{v}_{t,f},\bar{v}'_{t,a}\right),\quad\underline{r}^*_{t,u}:=r^*_u\left(\underline{v}_{t,f},\underline{v}'_{t,a}\right),$$
$$\bar{r}_{t,b}^*:=r_b^*\left(\bar{v}_{t,b}\right),\quad\underline{r}_{t,b}^*:=r^*_b\left(\underline{v}_{t,b}\right),$$

where $$\bar{v}_{t,f}:=\max_{\theta\in\Theta_{t,f}} x_t^\top \theta,\quad \underline{v}_{t,f}:=\min_{\theta\in\Theta_{t,f}} x_t^\top \theta, $$
$$\bar{v}_{t,b}:=\max_{\theta\in\Theta_{t,b}} x_t^\top \theta,\quad \underline{v}_{t,b}:=\min_{\theta\in\Theta_{t,b}} x_t^\top \theta, $$
$$\bar{v}'_{t,a}:=\bar{v}_{t,b}-\underline{v}_{t,f},\quad \underline{v}'_{t,a}:=\underline{v}_{t,b}- \bar{v}_{t,f}. $$

}
\IF{$\underline{r}^*_{t,b}>\bar{r}^*_{t,u}$}
\STATE{Choose bundling, i.e., $A_t=b$.}
\ELSIF{$\underline{r}^*_{t,u}>\bar{r}^*_{t,b}$}
\STATE{Choose unbundling, i.e., $A_t=u$.}
\ELSE
\STATE{Choose the strategy by
\begin{equation}
\label{IDS_exploration}
   A_t=\begin{cases}
 u &\text{if } \|x_t\|_{\Sigma^{-1}_{t-1,f}}\geq \|x_t\|_{\Sigma^{-1}_{t-1,b}}, \\
 b &\text{if } \|x_t\|_{\Sigma^{-1}_{t-1,b}}> \|x_t\|_{\Sigma^{-1}_{t-1,f}}.
\end{cases}
\end{equation}}

\ENDIF
\STATE{\textcolor{blue}{\%\% Set the price}}
\IF{$A_t=u$}
\STATE{\textcolor{blue}{\%\% Follow Algorithm~\ref{alg_pure_b}}}
\STATE{ Compute the LCB and UCB of the customer valuation:
$$\underline{v}_{t,f}=\min_{\theta\in\Theta_{t,f}} x_t^\top \theta,\quad \bar{v}_{t,a}=\max_{\theta\in\Theta_{t,a}} x_t^\top \theta, $$
}
\STATE{Set the prices by
$$
    p_{t,f}=p^*_f\left(\underline{v}_{t,f},r_a^*(\bar{v}_{t,a})\right)
$$
$$
   p_{t,a}=p_a^*\left(x_t^\top \hat{\theta}_{t,a}\right)
$$

}
\ELSE
\STATE{\textcolor{blue}{\%\% Follow certainty equivalent policy:}}
\STATE{Set the price by $p_{t,b}=p^*_b\left(x_t^\top \hat{\theta}_{t,b}\right)$}
\ENDIF

\ENDWHILE
\end{algorithmic}
\end{algorithm}

\begin{remark}
The choice of (\ref{IDS_exploration}) implements the idea of \textit{information directed sampling} (\citet{russo2018learning}). We will see shortly that the regret induced by a wrong choice (of sub-optimal selling strategy) is bounded by $O\left(\|x_t\|_{\Sigma^{-1}_{t-1,f}}+\|x_t\|_{\Sigma^{-1}_{t-1,b}}\right)$ for choosing either unbundling or bundling. Thus, the algorithm chooses the strategy that provides more information (for further periods)
and also guarantees an \textit{information ratio} bounded by constant. In fact, it is easy to check that UCB-based algorithm (for example, always choosing the selling strategy
with larger UCB of expected revenue) will not guarantee such a constant information ratio bound.
\end{remark}

\begin{comment}
\paragraph{Necessity of bundling to estimate $\theta_a$.}  We now construct a case where the seller will lose $O(T)$ revenue if only unbundling is used for exploring $\theta_a$ and $\theta_b$. Suppose $\epsilon_{t,a}=0$ for all $t$ and $\epsilon_{t,f}$ are i.i.d distributed by $F(v)=1-\frac{2}{2+v}$ for $v\geq 0$ and $F(v)=0$ when $v<0$. Suppose $X_t=1$ for all $t$ and $\theta^*_f=0$, $\theta^*_a=3$. The price range $p_t\in[0,T]$. Thus, the optimal expected revenue of unbundling is $3$ with $p_a=3$ and $p_f=0$. Also the optimal expected revenue of bundling is $3$ with $p_b=3$.

Now at the beginning, assume the seller knows $\theta^*_f=0$ and believes $\theta_a=0$, then the optimal prices under this belief are $p_f=p_b=T$, $p_a=0$, with expected revenues $\frac{2T}{2+T}$ and $\frac{2T}{T-1}$, which are both smaller than $3$ when $T>3$. In addition, at unbundling period, the probability of no ancillary item's observation is $1-\frac{2}{2+T}$. By noting that the probability of no ancillary item's observation (during unbundling periods) over whole horizon $T$ is lower bounded by $$\left(1-\frac{2}{2+T}\right)^T\geq \left(1-\frac{2}{2+T}\right)^{T+2}\geq \frac{1}{27},$$
If the seller refuses to use $\theta_b$ to estimate $\theta_a$, even he knows $\theta_f^*=0$, with probability at least $\frac{1}{27}$ he will always believe $\theta_a=0$ and will lose $O(T)$ revenues in total.
\end{comment}

\subsection{Regret Analysis}

The following theorem gives the regret bound for Algorithm~\ref{Alg2}. The regret bound consists of several parts: (i) the regret under the ``bad'' event when the confidence set does not cover the true parameter; (ii) the strategy regret when a sub-optimal selling strategy is chosen; (iii) the pricing regret induced by the sub-optimal price when restricted to one selling strategy.

\begin{theorem}
Under Assumption~\ref{assp_add},~\ref{assp_bound},~\ref{assp_distr} and with the regularization parameter $\lambda= 1$, the regret of Algorithm~\ref{Alg2} is bounded by
\begin{multline*}
  O\left(d\sqrt{T}\log T\right) =  \underbrace{6\bar{p}}_{\text{Regret under ``bad'' event}}+
    \underbrace{24\bar{\beta}\sqrt{dT\log\left(\frac{d+T}{d}\right)}}_{\text{Strategy regret (Lemma~\ref{reg_gen_choice_sum})}}\\
    +\underbrace{6\sqrt{2}\bar{\beta}\sqrt{dT\log\left(\frac{d+T}{d}\right)}+ 2d\eta\bar{\beta}^2 \log\left(\frac{d+T}{d}\right)}_{\text{Pricing regret when unbundling (Lemma~\ref{Unbund_price_reg})}}
    +\underbrace{2d\eta\bar{\beta}^2\log\left(\frac{d+T}{d}\right)
}_{\text{Pricing regret when bundling (Lemma~\ref{Reg_price_bundle})}},
\end{multline*}
 where $\eta=B+\bar{p}B'$ and $\bar{\beta}=2\bar{\theta}+ \frac{2\mu}{\nu}\sqrt{2\log\left(T\right)+d\log\left(\frac{d +T}{d}\right)}$.
\end{theorem}

In the following, we elaborate on these parts one by one.

\paragraph{Good event:}\

First, we consider the ``good'' event where the confidence sets cover the true parameters. As we use the same parameter estimation routine as the previous section, the proof of the following lemma is the same as that of Lemma~\ref{lemma_high_prob}.

\begin{lemma}[Good Event]
Let $$\mathcal{E}_{f}=\left\{\theta_{f}^*\in \Theta_{t,f} \text{ for } t=1,...,T\right\},$$
$$\mathcal{E}_{a}=\left\{\theta_{a}^*\in \Theta_{t,a} \text{ for } t=1,...,T \right\},$$
$$\mathcal{E}_{b}=\left\{\theta_{b}^*\in \Theta_{t,b} \text{ for } t=1,...,T \right\}.$$
We have
$$\prob(\mathcal{E}_{f} \cap \mathcal{E}_{a} \cap \mathcal{E}_b) \ge 1 -3/T.$$
\label{lemma_high_prob_1}
\end{lemma}

\paragraph{Strategy regret:}\

Next, we consider the \textit{strategy regret} which captures the regret induced by a sub-optimal selling strategy. That is, when the bundling (resp. unbundling) strategy is more profitable at time $t$, the selling strategy $A_t$ is chosen to be unbundling (resp. bundling). Specifically, the strategy regret is defined by
$$\mathrm{ChoiceReg} \coloneqq \sum_{t=1}^T \left\vert r^*_b\left(x_t^\top \theta_b^*\right)-r^*_u\left(x_t^\top \theta_f^*,x_t^\top \theta_a^*\right)\right\vert \cdot \mathbbm{1}_{A_t\neq A_t^*}$$
where $A_t^*$ denotes the optimal selling strategy at time $t$. We note that the revenues for both strategies are taken under the optimal prices, so the definition focuses exclusively on the revenue loss induced by the sub-optimal choice of the selling strategy.

\begin{lemma}
\label{reg_general_single}
Under the event $\mathcal{E}_{f}\cap \mathcal{E}_{b}$, the single-period strategy regret satisfies
$$\left\vert r^*_b\left(x_t^\top \theta_b^*\right)-r^*_u\left(x_t^\top \theta_f^*,x_t^\top \theta_a^*\right)\right\vert \cdot \mathbbm{1}_{A_t\neq A_t^*}\le 6\bar{\beta}\left(\|x_t\|_{\Sigma_{t-1,f}^{-1}}+\|x_t\|_{\Sigma_{t-1,b}^{-1}}\right)\cdot \mathbbm{1}_{A_t\neq A_t^*}.$$
\end{lemma}

Lemma~\ref{reg_general_single} establishes an upper bound for single-period strategy regret. Intuitively, $\|x_t\|_{\Sigma_{t-1,f}^{-1}}$ and $\|x_t\|_{\Sigma_{t-1,b}^{-1}}$ represent the sizes of the confidence sets. Under the good event, $A_t\neq A_t^*$ happens only when the decision rule \eqref{IDS_exploration} is adopted. Consequently, the two confidence sets will overlap with each other and the revenue gap can be bounded by the summation of two confidence sets.

\begin{lemma}
\label{reg_gen_choice_sum}
Under event $\mathcal{E}_f\cap \mathcal{E}_b$ and with any regularization parameter $\lambda\geq 1$, the total expected strategy regret satisfies
$$ \E_{\epsilon}\left[\sum_{t=1}^T \left\vert r^*_b\left(x_t^\top \theta_b^*\right)-r^*_u\left(x_t^\top \theta_f^*,x_t^\top \theta_a^*\right)\right\vert \cdot \mathbbm{1}_{A_t\neq A_t^*}\right] \le 24\bar{\beta}\sqrt{dT\log\left(\frac{d\lambda+T}{d\lambda}\right)},$$
where the expectation is taken with respect to the randomness in selling strategy $A_t$ induced by the randomness in utility shocks.
\end{lemma}

Lemma~\ref{reg_gen_choice_sum} builds upon Lemma~\ref{reg_general_single} and provides an upper bound for the expected strategy regret. The bound relies critically on the decision rule \eqref{IDS_exploration}. Specifically, the decision rule aligns the right-hand-side upper bound with the choice of the selling strategy and thus makes the Elliptical Potential Lemma applicable (Lemma~\ref{EPL_lemma}). Other than this point, the rest of the proof for Lemma~\ref{reg_gen_choice_sum} is similar to the previous case of pure unbundling.

\paragraph{Pricing regret:}

While the previous part analyzes the regret induced by a sub-optimal selling strategy, the pricing regret refers to that with a fixed strategy, the regret caused by a sub-optimal price. The following two lemmas state the pricing regret for unbundling and bundling time periods, respectively. Their analyses are similar to that of Theorem~\ref{UB_purepricing_2}.

\begin{lemma}
\label{Unbund_price_reg}
The expected pricing regret happened at unbundling periods (when $A_t= u$), under $\mathcal{E}_f\cap \mathcal{E}_a$ with any $\lambda\geq 1$,  is upper bounded by
$$6\sqrt{2}\bar{\beta}\sqrt{dT\log\left(\frac{d\lambda+T}{d\lambda}\right)}+ 2d\eta\bar{\beta} \log\left(\frac{d\lambda+T}{d\lambda}\right).$$
\end{lemma}

\begin{lemma}
\label{Reg_price_bundle}
The expected pricing regret happened at bundling periods (when $A_t= b$), under $\mathcal{E}_b$ with any $\lambda\geq 1$,  is upper bounded by
$$2d\eta\bar{\beta}^2\log\left(\frac{d\lambda+T}{d\lambda}\right).$$
\end{lemma}

\section{One-Switch Setting}\label{sec_one_switch}

In this section, we consider a setting where the seller may change the selling strategy at most once throughout the horizon. Specifically, we study a policy that first adopts the unbundling strategy and then may switch to the bundling strategy at a certain time point. Under the unbundling strategy, the seller may collect observations and learn the customer's utility of both the focal and ancillary product. In this sense, the seller uses the unbundling strategy in the short-term to learn the market and then decide its long-term strategy based on the learning outcome. In certain practical applications such a one-switch policy may be preferable, compared to the policy studied in the previous section which may entail the selling strategy to change frequently and by customer.

In this section, we require the features $X_t$'s to be i.i.d. generated (Assumption~\ref{assp_X}). Such assumption is necessary from a technical viewpoint: with only one switch of the selling strategy allowed throughout the horizon, it ensures that the seller can infer the knowledge about future arrivals based on the past observations.

Under this setting, the regret of a policy/algorithm $\pi$  is defined as
\begin{equation*}
\text{Reg}_{T}^{\pi} \coloneqq T\cdot \max\left\{\mathbb{E}_{X} \left[r^*_u\left(X^\top \theta_f^*,X^\top \theta_a^*\right)\right], \mathbb{E}_{X}\left[r^*_b\left(X^\top \theta_b^*\right)\right]\right\} -   \mathbb{E}\left[\sum_{t=1}^T R^{\pi}_t\right]
\end{equation*}
where the first two expectations are taken with respect to the covariates and the last expectation is  with respect to both covariates and potential randomness in policy $\pi$. Compared to the regret definition in the last section, the benchmark (the first summation in above) sticks to one selling strategy and picks the more profitable one among these two. For the second summation, $R^{\pi}_t$ denotes the revenue collected under the online policy $\pi.$

\subsection{One-Switch Algorithm}

Our one-switch Algorithm~\ref{alg:one_switch}  begins by executing the unbundling strategy until that it finds the bundling strategy is more profitable (when $\underline{r}^*_{t,b}\geq \bar{r}^*_{t,u}$). The difference between this and the previous personalized case is that in computing the upper and lower confidence bounds of the two strategies, we are comparing their long-term performance (when we stick to one strategy in long run) rather than the advantage of one strategy in a specific time period. Specifically, the upper and lower revenue bounds for the two strategies are defined by
\begin{equation}
\label{one_switch_UCB}
    \bar{r}_{t,u}^*,\underline{r}^*_{t,u}:=\frac{1}{t}\sum_{t'=1}^tr^*_u\left(X_{t'}^\top \hat{\theta}_{t+1,f},X_{t'}^\top \hat{\theta}_{t+1,a} \right)\pm\left(4\Bar{p}\sqrt{\frac{\log T}{t}}+\frac{2\Bar{\beta}}{t}\sum_{t'=1}^t \|X_{t'}\|_{\Sigma^{-1}_{t'-1,a}}\right),
\end{equation}
\begin{equation}
    \label{one_switch_UCB_2}
    \bar{r}_{t,b}^*,\underline{r}^*_{t,b}:=\frac{1}{t}\sum_{t'=1}^tr^*_b\left(X_{t'}^\top (\hat{\theta}_{t+1,f}+\hat{\theta}_{t+1,a}) \right)\pm\left(4\Bar{p}\sqrt{\frac{\log T}{t}}+\frac{2\Bar{\beta}}{t}\sum_{t'=1}^t \|X_{t'}\|_{\Sigma^{-1}_{t'-1,a}}\right).
\end{equation}

The first part in both definitions estimate the long-term revenue under each selling strategy, and the second part captures the size of the confidence set.

%In the first phase of Algorithm~\ref{alg:one_switch}, the algorithm performs exact same pricing policy as Algorithm~\ref{alg_pure_b}.
As in the case of Algorithm~\ref{alg_pure_b}, the unbundling strategy provides observations for the learning of both the focal and ancillary product. Once it accumulates enough samples and becomes confident that the bundling strategy is better, the algorithm switches to the bundling strategy for the remaining time periods.

\begin{algorithm}[ht!]
\caption{One-Switch Algorithm}
\label{alg:one_switch}
\begin{algorithmic}
\STATE{Tuning Parameter: Regularization parameter $\lambda$.}
\STATE{Initialize $A_1=A_2=\cdots = A_T=u$}
\WHILE{$t=1,...,T$}
\STATE{Observe $X_t=x_t$}
\IF{$A_t=u$}
\STATE{\textcolor{blue}{\%\% Follow Algorithm 1}}
\STATE{Compute the UCB and LCB of the customer valuation (notations same as in \S\ref{sec_pure_unbundle}):
$$\underline{v}_{t,f}=\min_{\theta\in\Theta_{t,f}} x_t^\top \theta,\quad \bar{v}_{t,a}=\max_{\theta\in\Theta_{t,a}} x_t^\top \theta, $$
}
\STATE{Set the price by
$$
    p_{t,f}=p^*_f\left(\underline{v}_{t,f},r_a^*(\bar{v}_{t,a})\right)
$$
$$
   p_{t,a}=p_a^*\left(x_t^\top \hat{\theta}_{t,a}\right)
$$
}
\ELSIF{$A_t = b$}
\STATE{\textcolor{blue}{\%\% Certainty-equivalent pricing}}
\STATE{Set the price by certainty equivalent (notations same as in \S\ref{sec_general}) $p_{t,b}=p^*_b\left(x_t^\top \hat{\theta}_{t,b}\right)$}
\ENDIF
\STATE{ Compute UCBs and LCBs of average revenues by \eqref{one_switch_UCB} and \eqref{one_switch_UCB_2}}
\IF{$\underline{r}^*_{t,b}\geq \bar{r}^*_{t,u}$}
\STATE Set $A_{t+1}=\cdots=A_T=b$ (Switch the strategy to bundling)
\ENDIF
\ENDWHILE
\end{algorithmic}
\end{algorithm}

\subsection{Regret Analysis}

Theorem~\ref{thm_one_switch} gives the regret of Algorithm~\ref{alg:one_switch}.
%The regret is decomposed into several parts.
The first part concerns the regret caused by the revenue confidence sets in \eqref{one_switch_UCB} and \eqref{one_switch_UCB_2}. The second part is the strategy regret induced by the initial unbundling exploration period; specifically, the strategy regret bounds the revenue loss when the optimal selling strategy is bundling but the algorithm starts with unbundling strategy. The remaining two parts concern the revenue loss induced by the pricing decisions. Specifically, after a number of time periods, the algorithm identifies the optimal selling strategy with high probability, and then the revenue loss in the remaining time periods is due to the pricing decisions. We defer the detailed analysis to Appendix~\S\ref{sec_proof_one_switch}.

\begin{theorem}
Under Assumption~\ref{assp_add},~\ref{assp_bound},~\ref{assp_distr},~\ref{assp_X} and with the regularization parameter $\lambda= 1$, the regret of Algorithm~\ref{alg:one_switch} is bounded by
\begin{multline*}
   O\left(d\log T \sqrt{\frac{T}{q^*}}\right) = \underbrace{22\bar{p}}_{\text{Regret under ``bad'' event}}
    +\underbrace{16\Bar{p}\sqrt{T\log T }+8\Bar{p}\sqrt{\frac{2dT}{q^*}\log\left(\frac{d+T}{d}\right) }}_{\text{Strategy regret}}\\
    +\underbrace{6\sqrt{2}\bar{\beta}\sqrt{dT\log\left(\frac{d+T}{d}\right)}+ 2d\eta\bar{\beta}^2 \log\left(\frac{d+T}{d}\right)}_{\text{Pricing regret at unbundling (Lemma~\ref{Unbund_price_reg})}}
    +\underbrace{2d\eta\bar{\beta}^2\log\left(\frac{d+T}{d}\right)
}_{\text{Pricing regret at bundling (Lemma~\ref{Reg_price_bundle})}},
\end{multline*}
where $\eta=B+\bar{p}B'$ and $\bar{\beta}=2\bar{\theta}+ \frac{2\mu}{\nu}\sqrt{2\log\left(T\right)+d\log\left(\frac{d +T}{d}\right)}$.
\label{thm_one_switch}
\end{theorem}

As in Theorem~\ref{UB_purepricing_2}, the regret bound involves the probability $q^*$ which governs the rate of observation samples from the ancillary product.

\section{Comments and Conclusion}
In this paper, we consider pricing and selling strategies for a focal-ancillary combination 
with unknown parameters in demand functions.   The dependence between the sequential purchase process creates novel
technical difficulties that we resolve.
Specifically, we design algorithms for pricing and choosing a selling mechanism with provable bounded 
regrets under three settings: (a) pure unbundling, (b) personalized mechanism, (c) 
initially unbundling and switch to bundling if necessary.

The key novelties in our analysis are the following: (a) We present two different ways 
to bound the single-period regret for pricing, depending on whether the focal product's purchase 
is observed or not. (b) To our knowledge, this is the first paper introducing the learning 
rate under optimal pricing in learning and pricing problem. (c) We apply 
\textit{Information Directed Sampling(IDS)} ideas of (\citet{russo2018learning}) 
to balance the learning and earning when deciding the selling strategy for exploration, 
which we believe is novel in the area of revenue management. 

One future direction of research is to extend to multiple ancillary products sold sequentially: to design sub-linear regret
algorithms to learn the parameters and price, as well as determine the optimal 
ordering of the ancillaries.

\bibliographystyle{informs2014}
\bibliography{bundle1}

\appendix

\section{Analysis of Regularized MLE Problem}

\renewcommand{\thesubsection}{A\arabic{subsection}}

\label{sec_MLE_analysis}

 Here, we provide a self-contained analysis of the regularized MLE problem, the backbone of all the three algorithms. The analysis largely mimics the analysis of bandits problem with a generalized linear dependence (\citet{filippi2010parametric}). To ease the notation burden, we omit all the subscripts for focal, ancillary or bundle ($f,a,b$), and write the true parameter as $\theta^*$ (which can be $\theta_f^*$, $\theta_a^*$ or $\theta_b^*$). The analyses of this section uses  Assumptions~\ref{assp_bound} and~\ref{assp_distr}.

Given data set $\mathcal{D}$, recall the likelihood function for parameter $\theta$ is defined by
\begin{equation}
\label{MLE}
    LL(\theta;\mathcal{D}) = \sum_{t\in \mathcal{T},d_{t}=1}\log(1-F_\epsilon(p_{t}-x_{t}^\top\theta ))+\sum_{t\in \mathcal{T},d_{t}=0}\log(F_\epsilon(p_{t}-x_{t}^\top\theta)).
\end{equation}
where $d_{t}$ is the realized demand under the true parameter $\theta^*$.

Let data set $\mathcal{D}_t=\{(p_{t'},x_{t'},d_{t'}),t'=1,..,t\}$ with $t\leq T$ and abbreviate the data set into the subscript by $LL_t(\theta)\coloneqq LL(\theta;\mathcal{D}_t)$. Define
\begin{equation}
\label{MLE_ridge}
    \hat{\theta}_t \coloneqq \argmin_{\theta\in \Theta}-LL_t(\theta)+\lambda\nu \|\theta\|_2^2,
\end{equation}
where $\lambda$ is  the regularization parameter and $\nu$ is defined following Assumption~\ref{assp_distr}.

The gradient and the Hessian are computed by
\begin{equation}
\nabla LL_t(\theta)=-\sum_{t'=1}^{t} \xi_{t'}(\theta)x_{t'}, \quad \nabla^2 LL_t (\theta)=-\sum_{t'=1}^{t}\eta_{t'}(\theta)x_{t'}x_{t'}^\top.
\end{equation}
%where $\nabla$ and $\nabla^2$ stands for the corresponding operator and the summand
$$
\xi_{t'}(\theta)\coloneqq -\log'F_\epsilon\left(p_{t'}- x_{t'}^\top\theta\right)\mathbbm{1}_{\{d_{t'}=0\}}-\log'\left(1-F_\epsilon(p_{t'}- x_{t'}^\top\theta)\right)\mathbbm{1}_{\{d_{t'}=1\}},
$$
$$
\eta_{t'}(\theta)\coloneqq -\log''F_\epsilon\left(p_{t'}- x_{t'}^\top\theta)\right)\mathbbm{1}_{\{d_{t'}=0\}}-\log''\left(1-F_\epsilon(p_{t'}- x_{t'}^\top\theta))\right)\mathbbm{1}_{\{d_{t'}=1\}}.
$$

The following lemma states that under a non-anticipatory policy, the sequence of $\{\xi_{t'}(\theta^*)\}_{t'=1}^{t}$ is a martingale difference sequence adapted to history observations with (zero-mean) $\mu^2$-sub-Gaussian increments.

\begin{lemma}
\label{}
Let $\mathcal{H}_{t}=\sigma \left(p_1,x_1,d_1,...,p_{t},x_{t},d_{t}\right)$ and $\mathcal{H}_{0}=\sigma\left(\emptyset, \Omega\right)$. For all $t=1,...,T$, we have
$$\E\left[\xi_{t}(\theta^*)|\mathcal{H}_{t-1}\right]=0.$$
In addition, $\xi_{t}(\theta^*)$ is sub-Gaussian with variance $\mu^2$ where $\mu$ is as following Assumption~\ref{assp_distr}.
\end{lemma}
\proof{} \noindent
We have
\begin{align*}
&\mathbb{E}\left[\xi_{t}(\theta^*)|\mathcal{H}_{t-1}\right]\\
&=-\frac{f_\epsilon(p_{t}- x_{t}^\top\theta^*)}{F_\epsilon(p_{t}- x_{t}^\top\theta^*)}\mathbb{E}\left[\mathbbm{1}_{\{d_{t}=0\}}|\mathcal{H}_{t-1}\right]+\frac{f_\epsilon(p_{t}- x_{t}^\top\theta^*)}{1-F_\epsilon(p_{t}- x_{t}^\top\theta^*)}\mathbb{E}\left[\mathbbm{1}_{\{d_{t}=1\}}|\mathcal{H}_{t-1}\right]\\
&=-\frac{f_\epsilon(p_{t}- x_{t}^\top\theta^*)}{F_\epsilon(p_{t}- x_{t}^\top\theta^*)}F_\epsilon(p_{t}- x_{t}^\top\theta^*)+\frac{f_\epsilon(p_{t}- x_{t}^\top\theta^*)}{1-F_\epsilon(p_{t}- x_{t}^\top\theta^*)}(1-F_\epsilon(p_{t}- x_{t}^\top\theta^*))\\
&=0.
\end{align*}
From Assumption $\ref{assp_bound}$,
$$\left|p_{t}- x_{t}^\top\theta^*\right|\leq \bar{p}+\bar{\theta},$$
and by its definition, $$\mu=\sup_{|v|\leq \bar{p}+\bar{\theta}}\{\max\{-\log'F_\epsilon(v),-\log'(1-F_\epsilon(v))\}\},$$
then
$$\xi_{t}(\theta^*)\leq \mu \quad a.s.,$$
which implies the sub-Gaussian parameter $\mu^2$.
\qedwhite

Let
$$S_{t}\coloneqq \sum_{t'=1}^{t}\frac{\xi_{t'}(\theta^*)}{\mu}x_{t'},$$
and recall the design matrix
$$\Sigma_{t}=\lambda I+\sum_{t'=1}^{t}x_{t'}x_{t'}^\top.$$

The following theorem measures $S_t$'s deviation in terms of a metric induced by $\Sigma_t.$ We note that the original statement in \citet{lattimore2020bandit} is stronger where the event holds for all $t\in \mathbb{N}$. Another specialized version of the statement that replaces the set $\{1,...,T\}$ with any of its subsets $\mathcal{T}$ will be frequently used in our analysis, where $\mathcal{T}$ denotes the time periods used for the estimation of a certain parameter.

\begin{theorem}[Theorem 20.4, \citet{lattimore2020bandit}]
\label{Lattimore}

For all $\lambda>0$ and $\delta\in(0,1)$,

$$\mathbb{P}\left( \exists t\in \{1...,T\}:  \left\|S_{t}\right\|^2_{\Sigma_{t}^{-1}}\geq 2\log\left(\frac{1}{\delta}\right)+\log\left(\frac{\det \Sigma_{t}}{\lambda^d}\right)\right)\leq \delta. $$
\end{theorem}

Then we can use the theorem to produce the following lemma on the estimation error of $\hat{\theta}.$ The proof is a combination of the linear regression analysis (with the application of above theorem) in \citet{lattimore2020bandit} and the MLE analysis in \citet{javanmard2019dynamic}.

\begin{lemma}
For any regularization parameter $\lambda>0$, the following bound holds
$$\mathbb{P}\left( \exists t\in \{1,...,T\}:  \left\|\hat{\theta}_t-\theta^* \right\|_{\Sigma_t}\geq2\sqrt{\lambda}\bar{\theta}+ \frac{2\mu}{\nu}\sqrt{2\log\left(\frac{1}{\delta}\right)+\log\left(\frac{\det \Sigma_t}{\lambda^d}\right)}\right)\leq \delta$$
for any $\delta\in(0,1)$.
\end{lemma}

\proof{} \noindent
We perform a second-order Taylor's expansion for the objective function of regularized MLE \eqref{MLE_ridge} around the true parameter $\theta^*$, we have
\begin{align}
    \label{Taylor}
    &-LL_t(\theta^*)+\lambda\nu\|\theta^*\|^2_2+LL_t(\theta)-\lambda\nu\|\theta\|^2_2\\
    &=-\left\langle -\nabla LL_t(\theta^*)+\lambda \nu\theta^*,\theta-\theta^*\right\rangle-\frac{1}{2}\left\langle\theta-\theta^*,\left(-\nabla^2 LL_t(\tilde{\theta})+\lambda\nu I\right)(\theta-\theta^*)\right\rangle \nonumber
\end{align}
for some $\tilde{\theta}$ on the line segment between $\theta$ and $\theta^*$.

From Assumption $\ref{assp_bound}$,
 $$|p_{t'}- x_{t'}^\top\theta|\leq |p_{t'}|+\|x_{t'}\|_{2}\|\theta\|_2\leq \bar{p}+\bar{\theta}.$$
By definition of $\nu$, we know $\eta_t(\tilde{\theta})\geq \nu$. Thus,
\begin{equation}
\label{grad2ine}
    -\nabla^2 LL_t(\tilde{\theta})\geq \nu\cdot \sum_{t'=1}^{t}x_{t'}x_{t'}^\top.
\end{equation}

Further, by the optimality of $\hat{\theta}_t$,
$$-LL_t(\theta^*)+\lambda\nu\|\theta^*\|^2_2\geq -LL_t (\hat{\theta}_t)+\lambda\nu \|\hat{\theta}_t\|^2_2.$$

From (\ref{Taylor}), we have
\begin{equation*}
\left\langle-\nabla LL_t(\theta^*)+\lambda \nu \theta^*,\hat{\theta}_t-\theta^*\right\rangle+\frac{1}{2}\left\langle\hat{\theta}_t-\theta^*,\left(-\nabla^2 LL_t(\tilde{\theta})+\lambda\nu I\right)(\hat{\theta}_t-\theta^*)\right\rangle \leq0.
\end{equation*}

Further, with Cauchy-Schwartz inequality and plugging (\ref{grad2ine}) in the above inequality,
\begin{align*}
    \left\|-\nabla LL_t(\theta^*)+\lambda \nu \theta^*\right \|_{\Sigma_t^{-1}}\left\|\hat{\theta}_t-\theta^*\right\|_{\Sigma_t} \geq& \left\langle -\nabla LL_t(\theta^*)+\lambda \nu\theta^*,\theta^*-\hat{\theta}_t\right\rangle\\
    \geq &\frac{1}{2}\left\langle\hat{\theta}_t-\theta^*,\left(-\nabla^2 LL_t(\tilde{\theta})+\lambda\nu I\right)(\hat{\theta}_t-\theta^*)\right\rangle \\
    \geq & \frac{\nu}{2}\left\langle\hat{\theta}_t-\theta^*,\Sigma_t (\hat{\theta}_t-\theta^*)\right\rangle\\
    = & \frac{\nu}{2} \left\|\hat{\theta}_t-\theta^*\right\|_{\Sigma_t}^2
\end{align*}
almost surely. Consequently,
\begin{equation}
    \left\|-\nabla LL_t(\theta^*)+\lambda \nu\theta^*\right \|_{\Sigma_t^{-1}}\geq \frac{\nu}{2} \left\|\hat{\theta}-\theta^*\right\|_{\Sigma_t} \quad a.s.
    \label{tmp_taylor}
\end{equation}

Recall that
$$S_{t}=\sum_{t'=1}^{t}\frac{\xi_{t'}(\theta^*)}{\mu}x_{t'}=-\frac{1}{\mu}\nabla LL_t(\theta^*),$$
which implies
\begin{align*}
    \frac{\nu}{2\mu} \left\|\hat{\theta}_t-\theta^*\right\|_{\Sigma_t}
    \leq& \frac{1}{\mu}\left\|-\nabla LL_t(\theta^*)+\lambda \nu\theta^*\right  \|_{\Sigma_t^{-1}} \\
    = &\frac{1}{\mu}\left\|\mu S_t+\lambda \nu \theta^*\right \|_{\Sigma_t^{-1}}\\
    \leq& \left\|S_t\right\|_{\Sigma^{-1}_{t}}+\frac{\sqrt{\lambda}\nu}{\mu}\sqrt{(\theta^*)^\top (\lambda \Sigma_{t}^{-1})\theta^*}\\
    \leq& \left\|S_t\right\|_{\Sigma^{-1}_{t}}+\frac{\sqrt{\lambda}\nu}{\mu}\|\theta^*\|_2.
\end{align*}
Here the first line comes from \eqref{tmp_taylor}, the second line comes from the definition of $S_t$, the third line comes from the norm inequality, and the last line is from the fact that $\lambda \Sigma_{t}^{-1} \le I$. Thus, we complete the proof from combining Theorem (\ref{Lattimore}) with $\|\theta^*\|_2\leq \bar{\theta}$.
\qedwhite

Choose $\delta=\frac{1}{T}$, and recall the definition
\begin{equation*}
  \beta(\Sigma)=2\sqrt{\lambda}\bar{\theta}+ \frac{2\mu}{\nu}\sqrt{2\log T+\log\left(\frac{\det \Sigma}{\lambda^d}\right)}.
\end{equation*}
We obtain the following corollary.
\begin{corollary}
\label{MLEbound}
For all $\lambda>0$,
$$\mathbb{P}\left( \exists t\in \{1,...,T\}:  \left\|\hat{\theta}_t-\theta^* \right\|_{\Sigma_t}\geq \beta(\Sigma_t) \right)\leq \frac{1}{T}. $$
\end{corollary}

Lastly, we introduce a uniform upper bound of $\beta(\Sigma_t)$'s. Note that $\|x_{t}\|_2^2\leq 1$ by assumption, we can apply Lemma 19.4 of \citet{lattimore2020bandit} (purely algebraic analysis with no stochasticity) and obtain the following corollary.
\begin{corollary}
\label{ub_beta}
For all $t=1,...,T$,
$$\beta(\Sigma_t)\leq \bar{\beta}:=2\sqrt{\lambda}\bar{\theta}+ \frac{2\mu}{\nu}\sqrt{2\log T+d\log\left(\frac{d\lambda +T}{d\lambda}\right)}.$$
\end{corollary}

\section{Proof of Section~\ref{sec_pure_unbundle}}

\label{sec_proof_pure_unb}

\renewcommand{\thesubsection}{B\arabic{subsection}}

\subsection{Proof of Lemma~\ref{price_1}}
\proof{}\noindent
We first argue that $p_f^*(v_f,r^*_a(v_a))$ is strictly increasing in $v_f$ and strictly decreasing in $v_a$.
Recall the optimal pricing function is
$$p^*_f(v_f,r_a)=g_f(v_f+r_a)-r_a.$$

By $g'(v)\in (0,1)$ from Lemma~\ref{Jarlemma_2} in the following, we have $p_f^*(v_f,r_a)$ is strictly increasing in $v_f$ and strictly decreasing in $r_a$. Further, since
$$r^{*'}_a(v_a)=1-F_{\epsilon_a}(g_a(v_a)-v_a)>0,$$
we also have $r^*_a(\cdot)$ is strictly increasing.

Given the event $\mathcal{E}_f\cap \mathcal{E}_a$, we have $\theta^*_f\in\Theta_{t,f}$ and $\theta^*_a\in \Theta_{t,a}$. By the definition of $\underline{v}_{t,f}$ and $\bar{v}_{t,a}$, we complete the proof.
\qedwhite

\begin{lemma}[Lemma 14 in \citet{javanmard2019dynamic}]
\label{Jarlemma_2}
If $1-F_\epsilon$ is log-concave, then the price function $g$ satisfies $0<g'(v)<1$ for all values of $v\in \mathbb{R}$.
\end{lemma}

\subsection{A Few Additional Lemmas on the Revenue and Pricing Functions}

The following lemma analyzes the revenue gap and the optimal pricing gap for the ancillary product.

\begin{lemma}[\citet{javanmard2019dynamic} Section 8, page 22]
\label{Rev_1}
We have
$$r^*_a(v_{a})-r_a(p_a;v_a)\leq (B+\bar{p}B')|p_a-p^*_a(v_a)|^2,$$
$$|p_a^*(v)-p_a^*(v_{a})|\leq |v-v_{a}|$$
where $\bar{p}$ is defined in Assumption~\ref{assp_bound}, and $B$ and $B'$ are defined in Assumption~\ref{assp_distr}.
\end{lemma}

For its proof, note that from \citet{javanmard2019dynamic}
$$r_a^*(v_{a})-r_a(p_a;v_{a})\leq \left(B+\frac{\bar{p}}{2} \cdot \max_{v\in \left[\underline{p}-\bar{\theta}, \Bar{p}+\bar{\theta}\right]} f'_{\epsilon_a}(v)\right)|p_a-p^*_a(v_{a})|^2,$$
where we apply a looser bound $$\left(B+\frac{\bar{p}}{2} \cdot \max_{v\in  \left[\underline{p}-\bar{\theta}, \Bar{p}+\bar{\theta}\right]} f'_{\epsilon_a}(v)\right)|p_a-p^*_a(v_{a})|^2\leq  (B+\bar{p}B')|p_a-p_a^*(v_{a})|^2.$$

The following lemma fixes the ancillary price as optimal and analyzes the effect of the focal price $p_f.$

\begin{lemma}
We have
\label{rev_3}
    $$r_{u}^*(v_f,v_a)-r_u\left(p_f,p^*_a(v_a);v_f,v_a\right)
    \leq (B+\bar{p}B')|p_f-p_f^*(v_f,r_a^*(v_a))|^2$$
where $\bar{p}$ is defined in Assumption~\ref{assp_bound}, and $B$ and $B'$ are defined in Assumption~\ref{assp_distr}.
\end{lemma}
\proof{}\noindent
As in this case $p^*_a(v_a), v_f,v_a$ are fixed, we denote $r_{u,p_f}(p):=r_{u}(p,p^*_a(v_{a});v_{f},v_{a})$ for brevity. Then its first and second derivatives are
$$r'_{u,p_f}(p)=1-F_{\epsilon_f}(p-v_{f})-(r_a^*(v_{a})+p)f_{\epsilon_f}(p-v_{f}),$$
and
$$r''_{u,p_f}(p)=-2f_{\epsilon_f}(p-v_{f})-(r_a^*(v_{a})+p)f'_{\epsilon_f}(p-v_{f}).$$

Recall $0\leq r_a^*(v_a)\leq \bar{p}$, $0\leq p\leq \bar{p}$ and with corresponding definitions of $B$, $B'$, we have  $\left|r''_{u,p_f}(p)\right|\leq 2\bar{p}B'+2B$ for all $p$.

Also, from the optimality of $p_f^*(v_f,r^*_a(v_a))$, we know $r'_{u,p_f}\left(p_f^*(v_f,r^*_a(v_a))\right)=0$. Then we complete our proof by applying Talyor expansion.
\qedwhite

The following lemma concerns the smoothness of the ancillary revenue in terms of the ancillary price. Also, it states that the optimal price of the ancillary product is $1$-Lipschitz in its valuation.

\begin{lemma}[\citet{javanmard2019dynamic} Section 8, page 22]
\label{Rev_11}
We have
$$r^*_a(v_{a})-r_a(p_a;v_a)\leq (B+\bar{p}B')|p_a-p^*_a(v_a)|^2,$$
and
$$|p_a^*(v)-p_a^*(v_{a})|\leq |v-v_{a}|.$$
\end{lemma}

The following lemma extends Lemma~\ref{Rev_11} to show that the optimal ancillary revenue is $1$-Lipschitz of its valuation.

\begin{lemma} We have
\label{Rev_2}
$$|r_a^*(v_a)-r_a^*(v_a')|\leq  |v_a-v_a'|.$$
\end{lemma}

\proof{}\noindent
Recall that
$$r_a^*(v)=g_a(v)(1-F_{\epsilon_a}(g_a(v)-v)),$$
$$\frac{d r_a^*(v)}{d v}=g'_a(v)(1-F_{\epsilon_a}(g_a(v)-v))+(1-g'_a(v))g_a(v)f_{\epsilon_a}(g_a(v)-v)=1-F_{\epsilon_a}(g_a(v)-v),$$
where the last equality is by definition of $g_a(\cdot)$. Also, by Assumption~\ref{assp_distr}, the log-concavity of distributions imply $1-F_{\epsilon_a}(\cdot)\in (0,1)$ under the feasible domain.  Thus
$$0<\frac{d r_a^*(v)}{d v}< 1. $$
And by Taylor Expansion, we complete the proof.
\qedwhite

The following Lemmas express how the price and revenue of the joint (of the focal and ancillary) changes in terms of their valuation.

\begin{lemma}
\label{Rev_4}
We have
$$|p_f^*(v_f,r_a)-p_f^*(v'_f,r'_a)|\leq |v_f-v'_f|+2|r_a-r'_a|$$
where $p_f^*$ is the optimal pricing function (for focal product) defined in Section~\ref{sec_revenue_func}.
\end{lemma}

\proof{}\noindent
To see this,
\begin{align*}
    &|p_f^*(v_f,r_a)-p_f^*(v'_f,r'_a)|\\
    &= |g_f(v_f+r_a)-r_a-g_f(v'_f+r'_a)+r'_a|\\
    &\leq |g_f(v_f+r_a)-g_f(v'_f+r'_a)|+|r'_a-r_a|\\
    &\leq |v_f+r_a-(v'_f+r'_a)|+|r'_a-r_a| \quad \text{(Lemma~\ref{Jarlemma_2})}\\
    &\leq |v_f-v'_f|+2|r_a-r'_a|,
\end{align*}
\qedwhite

\begin{lemma}
\label{Revenue_price_con}
The following bound holds for all possible $v_f, v_f'$ and $v_a, v_a'$,
$$r_{u}^*\left(v_{f},v_{a}\right)-r_{u}\left(p_f^*(v'_f,r_a^*(v'_a)),p_a^*(v'_a);v_{f},v_{a}\right)\leq 9\eta\left(|v_{f}-v'_{f}|^2+|v_{a}-v'_{a}|^2\right)$$
where $\eta=B+\bar{p}B'$.
\end{lemma}

\proof{}\noindent
We have
\begin{align*}
   & r_{u}^*\left(v_{f},v_{a}\right)-r_{u}^*\left(v'_f,v'_a;v_{f},v_{a}\right)\\
   &= r^*_{u}\left(v_{f},v_{a}\right)-r_{u}\left(p_f^*(v'_{f},r_a^*(v'_{a})),p_a^*(v_a);v_{f},v_{a}\right)\\
   &  \ +
   r_{u}\left(p_f^*(v'_{f},r_a^*(v'_{a})),p_a^*(v_a);v_{f},v_{a}\right)-r_{u}\left(p_f^*(v'_{f},r_a^*(v'_{a})),p_a^*(v'_a);v_{f},v_{a}\right)\\
   &\leq r^*_{u}\left(v_{f},v_{a}\right)-r_{u}\left(p_f^*(v'_{f},r_a^*(v'_{a})),p_a^*(v_a);v_{f},v_{a}\right)+r_a^*(v_a)-r_a(p_a(v'_{a});v_{a})\\
   &\leq  r^*_{u}\left(v_{f},v_{a}\right)-r_{u}\left(p_f^*(v'_{f},r_a^*(v'_{a})),p_a^*(v_a);v_{f},v_{a}\right)+(B+\bar{p}B')|v'_{a}-v_{a}|^2 \quad \text{(Lemma~\ref{Rev_1})}\\
   &\leq (B+\bar{p}B')|p_f^*\left(v_{f},r_a^*(v_{a})\right)-p^*_f(v'_{f},r_a^*(v'_{a})) |^2+(B+\bar{p}B')|v'_{a}-v_{a}|^2 \quad \text{ (Lemma~\ref{rev_3})}\\
   &\leq  (B+\bar{p}B')\left(2|v'_{f}-v_f|^2 +8\left|r^*_a(v'_{a})-r^*_a\left(v_{a}\right)\right|^2+\left|v'_{a}-v_{a}\right|^2\right) \quad \text{ (Lemma~\ref{Rev_4})}\\
      &\leq  (B+\bar{p}B')\left(2|v'_{f}-v_f|^2 +8\left|v'_a-v_a\right|^2+\left|v'_{a}-v_{a}\right|^2\right)  \quad \text{ (Lemma~\ref{Rev_2})}\\
   &\leq 9(B+\bar{p}B')\left(|v_{f}-v'_{f}|^2+|v_{a}-v'_{a}|^2\right).
\end{align*}
\qedwhite

\begin{lemma}
\label{Revenue_con}
We have
 $$|r_{u}^*\left(v_{f}, v_{a}  \right)-r_{u}^*\left(v'_f, v'_a  \right)|
    \leq |v_f-v'_f|+|v_a-v'_a|.$$
\end{lemma}

\proof{}\noindent
Recall from the discussion in Section~\ref{sec_revenue_func},
$$r_{u}^*(v_f,v_a)=r_f^*(v_f+r_a^*(v_a)).$$
We can apply Lemma~\ref{Rev_2} twice for $r_f^*(\cdot)$ and $r_a^*(\cdot)$ respectively.
\begin{align*}
     &|r_{u}^*(v_f,v_a)  -r^*_{u}(v'_f,v'_a)|\\
    &\leq|v_{f}-v'_{f}+r_a^*(v_{a})-r^*_a(v'_{a})| \quad \text{ (Lemma~\ref{Rev_2})}\\
    &\leq|v_{f}-v'_{f}|+|v_{a}-v'_{a}|\quad \text{ (Lemma~\ref{Rev_2})}.
\end{align*}

\qedwhite

\subsection{Proofs of two Single-Period Regret Bounds}

\paragraph{Proof of Lemma~\ref{purepricing_singlereg_1}}

\proof{}\noindent
For brevity, we omit the period $t$ subscript in notations when it's clear. Let $\hat{v}_f=x^\top\hat{\theta}_{f}$, $v_f^*=x^\top\theta^*_{f}$ and $\hat{v}_a=x^\top\hat{\theta}_{a}$, $v_a^*=x^\top\theta^*_{a}.$ Also, let $\bar{p}_a=p_a^*(\bar{v}_a)$ be the optimal price given the ``most optimistic'' ancillary valuation. Under $\mathcal{E}_f\cap \mathcal{E}_a$,
\begin{align*}
&r_u^*\left(v^*_{f},v^*_{a}\right)-r_u\left(p_f,p_a;v^*_{f},v^*_{a}\right)\\
&=r_u^*\left(v^*_{f},v^*_{a}\right)-r_u\left(p_f,\bar{p}_a;v^*_f,v^*_a\right)+ r_u\left(p_f,\bar{p}_a;v^*_f,v^*_a\right)- r_u\left(p_f,p_a;v^*_f,v^*_a\right)\\
&\leq 9\eta\left(\left|v^*_f-\underline{v}_f\right|^2+\left|v^*_a-\bar{v}_a\right|^2\right) +r_u\left(p_f,\bar{p}_a;v^*_f,v^*_a\right)- r_u\left(p_f,p_a;v^*_f,v^*_a\right)\quad \text{(Lemma~\ref{Revenue_price_con})}\\
&\leq 9\eta\left(\left|v^*_f-\underline{v}_f\right|^2+\left|v^*_a-\bar{v}_a\right|^2\right) +r_a\left(\bar{p}_a;v^*_a\right)- r_a\left(p_a;v^*_a\right)\\
&\leq  9\eta\left(\left|v^*_f-\underline{v}_f\right|^2+\left|v^*_a-\bar{v}_a\right|^2\right) +r^*_a\left(v^*_a\right)- r_a\left(p_a;v^*_a\right) \quad \text{(By optimality of $r_a^*(V_a)$)}\\
&\leq  9\eta\left(\left|v^*_f-\underline{v}_f\right|^2+\left|v^*_a-\bar{v}_a\right|^2\right)+\eta\left|v^*_a-\hat{v}_a\right|^2 \quad \text{(Lemma~\ref{Rev_11})}\\
&\leq  9\eta\left(\left|v^*_f-\hat{v}_f+\hat{v}_f-\underline{v}_f\right|^2+\left|v^*_a-\hat{v}_a+\hat{v}_a-\bar{v}_a\right|^2+\left|v^*_a-\hat{v}_a\right|^2\right)\\
&\leq  36\eta\left(\left|v^*_f-\hat{v}_f\right|^2+\left|\hat{v}_f-\underline{v}_f\right|^2+\left|v^*_a-\hat{v}_a\right|^2+\left|\hat{v}_a-\bar{v}_a\right|^2\right)\\
&\leq 36\eta \left(\left\|x_t\right\|^2_{\Sigma_{t-1,f}^{-1}}\left\|\theta_f^*-\hat{\theta}_{t,f}\right\|^2_{\Sigma_{t-1,f}}+\left|\hat{v}_f-\underline{v}_f\right|^2\right)+36\eta\left(\|x_t\|^2_{\Sigma_{t-1,a}^{-1}}\left\|\theta_a^*-\hat{\theta}_{t,a}\right\|^2_{\Sigma_{t-1,a}}+\left|\hat{v}_a-\bar{v}_a\right|^2\right)\\
&\leq   72\eta \left(\|x_t\|^2_{\Sigma_{t-1,f}^{-1}}\beta^2(\Sigma_{t-1,f})+\|x_t\|^2_{\Sigma_{t-1,a}^{-1}}\beta^2(\Sigma_{t-1,a})\right) \quad \text{(By $\mathcal{E}_f\cap \mathcal{E}_a$)}\\
&\leq  72\eta \bar{\beta}^2 \left(\|x_t\|^2_{\Sigma_{t-1,f}^{-1}}+\|x_t\|^2_{\Sigma_{t-1,a}^{-1}}\right) \quad \text{(Using $\bar{\beta}\geq \beta(\Sigma_{t-1,f}),\beta(\Sigma_{t-1,a})$)}\\
&\leq 144\eta \bar{\beta}^2\|x_t\|^2_{\Sigma_{t-1,a}^{-1}}\quad \text{(Using $\Sigma_{t-1,f}\geq \Sigma_{t-1,a}$ )}.
\end{align*}
\qedwhite

\paragraph{Proof of Lemma~\ref{purepricing_singlereg_2}}

\proof{}\noindent
For brevity, we omit the period $t$ subscript in notations when it's clear. Denote $q_f^*:=1-F_{\epsilon_f}(p^*_{f}-v^*_{f})$, i.e., the focal item purchase probability under optimal pricing. With $\mathcal{E}_f\cap \mathcal{E}_a$,
\begin{align*}
&r_u^*(v^*_f,v^*_a)-r_u\left(p_f,p_a;v^*_f,v^*_a\right)\\
&=q^*_f\cdot \left(p_f^*+r_a^*\left(v^*_a\right)\right)- r_u\left(p_f,p_a;v^*_f,v^*_a\right)\\
&\leq q_f\cdot \left(p_f^*+r_a^*\left(v^*_a\right)\right)- r_u\left(p_f,p_a;v^*_f,v^*_a\right) \quad \text{(Lemma~\ref{price_1})}\\
&= q_f\cdot \left(p_f^*+r_a^*\left(v^*_a\right)\right)- q_f\left(p_f+r_a\left(p_a^*(\hat{v}_a);v^*_a\right)\right)\\
&=  q_f\cdot \left(p_f^*-p_f+r_a^*\left(v^*_a\right)-r_a\left(p_a^*(\hat{v}_a);v^*_a\right)\right)\\
&\leq  q_f\cdot \left(\left|v^*_f-\underline{v}_f\right|+2\left|r^*_a(\bar{v}_a)-r^*_a(v^*_a)\right|+r_a^*\left(v^*_a\right)-r_a^*\left(p_a^*(\hat{v}_a);v^*_a\right)\right) \quad \text{(Lemma~\ref{Rev_4})}\\
&\leq  q_f\cdot \left(\left|v^*_f-\underline{v}_f\right|+2\left|\bar{v}_a-v^*_a\right|+r_a^*\left(v^*_a\right)-r_a^*\left(p_a^*(\hat{v}_a);v^*_a\right)\right) \quad \text{(Lemma~\ref{Rev_2})}\\
&\leq  q_f\cdot \left(\left|v^*_f-\underline{v}_f\right|+2\left|\bar{v}_a-v^*_a\right|+\eta\left(v^*_a-\hat{v}_a\right)^2\right) \quad \text{(Lemma~\ref{Rev_11})}\\
&\leq q_f\left(\left\|x_t\right\|_{\Sigma_{t-1,f}^{-1}}\left\|\theta_f^*-\hat{\theta}_{t,f}\right\|_{\Sigma_{t-1,f}}+|\hat{v}_f-\underline{v}_f|\right)\\\
&\ +2q_f\left(\left\|x_t\right\|_{\Sigma_{t-1,a}^{-1}}\left\|\theta_a^*-\hat{\theta}_{t,a}\right\|_{\Sigma_{t-1,a}}+|\hat{v}_a-\bar{v}_a|\right)+q_f\eta\left\|x_t\right\|^2_{\Sigma_{t-1,a}^{-1}}\left\|\theta_a^*-\hat{\theta}_{t,a}\right\|^2_{\Sigma_{t-1,a}}\\
&\leq 2q_f\bar{\beta}\left\|x_t\right\|_{\Sigma_{t-1,f}^{-1}}+4q_f\bar{\beta}\left\|x_t\right\|_{\Sigma_{t-1,a}^{-1}}+q_f\eta\bar{\beta}^2\left\|x_t\right\|^2_{\Sigma_{t-1,a}^{-1}}\\
&\leq q_f\left(6\bar{\beta}\left\|x_t\right\|_{\Sigma_{t-1,a}^{-1}}+\eta \bar{\beta}^2\left\|x_t\right\|^2_{\Sigma_{t-1,a}^{-1}}\right) \quad \text{(Using $\Sigma_{t-1,f}\geq \Sigma_{t-1,a}$ )}.
\end{align*}
\qedwhite

\subsection{Proof for Theorem~\ref{UB_purepricing_2} and Theorem~\ref{UB_purepricing_1}}

\label{sec_UB_purepricing}

We first introduce the elliptical potential lemma, which turns out to be useful in our analysis. This lemma is first introduced by \citet{lai1982least} and then wildly used for proving the regret bound for stochastic linear bandit and its variants (See \citet{lattimore2020bandit}).
\begin{lemma}
\label{EPL_lemma}
For a constant $\lambda\geq 1$ and a sequence of $\{x_t\}_{t\geq 1}$ with $\|x_t\|_2\leq 1$ for all $t\geq 1$ and $x_t\in \mathbb{R}^d$, define the sequence of covariance matrices:
$$\Sigma_0:=\lambda I_d, \quad \Sigma_t:=\lambda I_d+\sum_{t'=1}^tx_{t'}x_{t'}^\top \ \ \forall t\geq 1,$$
where $I_d$ is the identity matrix with dimension $d$. Then for any $T\geq 1$, the following inequality holds
$$\sum_{t=1}^T\|x_{t}\|_{\Sigma^{-1}_{t-1}}\leq 2d\log\left(\frac{\lambda d+T}{\lambda d}\right).$$
\end{lemma}

Now we proceed to the proof of our theorems.

\textbf{Regret from ``bad'' event.} From Lemma~\ref{lemma_high_prob},
$$\prob\left(\mathcal{E}_f^c\cup \mathcal{E}_a^c\right)\leq \frac{2}{T}.$$
Under this bad event, it will cause at most $2\bar{p}T$ regrets by noticing the single period regret is upper bounded by $2\bar{p}$. Thus the total expected regret when $\mathcal{E}_f\cap \mathcal{E}_a$
does not happen is thus bounded by $2\bar{p}$.

%With a bit abuse of the notation, we assume the whole probability space to be $\mathcal{E}_f\cap \mathcal{E}_a$ and the probability and expectation are adjusted to the conditional ones (conditional on the event of $\mathcal{E}_f\cap \mathcal{E}_a$). This makes no essential change to the analysis provided the high probability of the event, but would greatly simplify the notation.

We define a counting process for the focal product purchase as follows. Let $t_0=0.$ For $k=1,2,...$, denote
$$t_k \coloneqq \min \left\{t\Bigg |\sum_{t'=1}^t \mathbbm{1}_{d_{t',f}=1} \ge k\right\}$$
as the first time that we have observed $k$ focal product purchase (i.e., effective ancillary samples). Denote the interarrival time as $\tau_{k}:=t_{k}-t_{k-1}$ and the corresponding counting process as $N(t)$, which is the count of focal purchases up to period $t$ (inclusive).

\paragraph{Regret analysis for Theorem~\ref{UB_purepricing_2}.}

From Lemma~\ref{purepricing_singlereg_2}, we know the cumulative regret has the following bound
\begin{align}
\sum_{t=1}^T\E[\mathrm{Reg}_t \cdot \mathbbm{1}_{\mathcal{E}_f \cap \mathcal{E}_a}]
    \leq \sum_{t=1}^T\E\left[q_{t,f}\left(6\bar{\beta}\left\|x_t\right\|_{\Sigma_{t-1,a}^{-1}}+\eta\bar{\beta}^2\left\|x_t\right\|_{\Sigma_{t-1,a}^{-1}}^2\right)\right] \label{reg_sum_new}.
\end{align}

Noting that $q_{t,f}=\E\left[\mathbbm{1}_{\{d_{t,f}=1\}}|x_t\right]$, we can follow the rule of conditional expectation and further express $(\ref{reg_sum_new})$ by
\begin{align}
   (\ref{reg_sum_new})=& \sum_{t=1}^T\E\left[ \E\left[\mathbbm{1}_{\{d_{t,f}=1\}}|x_t\right]\left(6\bar{\beta}\left\|x_t\right\|_{\Sigma_{t-1,a}^{-1}}+\eta\bar{\beta}^2\left\|x_t\right\|_{\Sigma_{t-1,a}^{-1}}^2\right)\right] \nonumber\\
   =& \E\left[\sum_{t=1}^T \mathbbm{1}_{\{d_{t,f}=1\}}\left(6\bar{\beta}\left\|x_t\right\|_{\Sigma_{t-1,a}^{-1}}+\eta\bar{\beta}^2\left\|x_t\right\|_{\Sigma_{t-1,a}^{-1}}^2\right)\right] \label{reg_sum_new_3}.
\end{align}

Note that $\sum_{t=1}^T \mathbbm{1}_{\{d_{t,f}=1\}}=N(T)$. The summation only involves periods when $d_{t,f}=1$, so $(\ref{reg_sum_new_3})$ can be further expressed as:
\begin{align*}
    (\ref{reg_sum_new_3})=&\E\left[\sum_{k=1}^{N(T)}\left(6\bar{\beta}\left\|x_{t_{k}}\right\|_{\Sigma_{t_{k}-1,a}^{-1}}+\eta\bar{\beta}^2\left\|x_{t_{k}}\right\|_{\Sigma_{t_{k}-1,a}^{-1}}^2\right)\right]\\
    \leq & \E\left[\left(6\bar{\beta}\sqrt{N(T)\sum_{k=1}^{N(T)}\left\|x_{t_{k}}\right\|_{\Sigma_{t_{k}-1,a}^{-1}}^2}+\eta\bar{\beta}^2\sum_{k=1}^{N(T)}\left\|x_{t_{k}}\right\|_{\Sigma_{t_{k}-1,a}^{-1}}^2\right)\right]
\end{align*}
where the second line applies Cauchy-Schwartz inequality.

Thus, with $N(T)\leq T$ almost surely, we can finally apply Lemma~\ref{EPL_lemma} and complete the proof.

\paragraph{Proof for Theorem~\ref{UB_purepricing_1}.}

From Lemma~\ref{purepricing_singlereg_1}, we know,
\begin{align}
    \sum_{t=1}^T\E[\mathrm{Reg}_t \cdot \mathbbm{1}_{\mathcal{E}_f \cap \mathcal{E}_a}] \le 144\eta \bar{\beta}^2\sum_{t=1}^T\E\left[\left\|X_t\right\|^2_{\Sigma_{t-1,a}^{-1}}\cdot \mathbbm{1}_{\mathcal{E}_f \cap \mathcal{E}_a}\right]. \label{reg_sum}
\end{align}
In this proof, we capitalize the covariate $X_t$ to emphasize that it is sampled i.i.d. from some distribution, and we will use $x_t$ to denote its realization. Also, for brevity, for any random variable $Z$, we use $\E_G[Z]=\E[Z\cdot \mathbbm{1}_{\mathcal{E}_f \cap \mathcal{E}_a}]$.  

The inequality reduces the regret to a summation with respect to only the ancillary product. The summation is still difficult to analyze in that  $\Sigma_{t,a}$ will only be updated when a focal purchase takes place, instead of for all the time periods with every $X_t$ as in linear bandits. As a result, we can not directly use the elliptical potential lemma (Lemma~\ref{EPL_lemma}). As the analysis only concerns ancillary $a$, we will omit the subscript $a$ in notations when the context is clear.

We can rearrange ($\ref{reg_sum}$) as
\begin{align*}
    &144\eta \bar{\beta}^2\sum_{t=1}^T\E_G\left[\left\|X_t\right\|^2_{\Sigma_{t-1}^{-1}}\right]\\
    &= 144\eta \bar{\beta}^2\E_G\left[\sum_{k=1}^{N(T)}\sum_{t=t_{k-1}+1}^{t_{k}} \left\|X_t\right\|^2_{\Sigma_{t-1}^{-1}}\right]+144\eta \bar{\beta}^2\E_G\left[\sum_{t=t_{N(T)}+1}^{T} \left\|X_t\right\|^2_{\Sigma_{t-1}^{-1}}\right]\\
    &= 144\eta \bar{\beta}^2\E_G\left[\sum_{k=1}^{N(T)}\sum_{t=t_{k-1}+1}^{t_{k}} \left\|X_t\right\|^2_{\Sigma_{t_{k-1}}^{-1}}\right]+144\eta \bar{\beta}^2\E_G\left[\sum_{t=t_{N(T)}+1}^{T} \left\|X_t\right\|^2_{\Sigma_{t_{N(T)}}^{-1}}\right],
\end{align*}
where the last equality is because the matrix $\Sigma_t$ is only updated at $t_{k}$ for some $k$.

\begin{align}
    &\E_G\left[\sum_{i=1}^{\tau_k} \left\|X_{t_{k-1}+i}\right\|^2_{\Sigma_{t_{k-1}}^{-1}} \bigg \vert \Sigma_{t_{k-1}} \right] \nonumber\\
    &= \E\left[ \E_G\left[ \sum_{i=1}^{\tau_k} \left\|X_{t_{k-1}+i}\right\|^2_{\Sigma_{t_{k-1}}^{-1}} \bigg \vert \Sigma_{t_{k-1}},\{X_{t'}\}_{t'=t_{k-1}+1}^{T} \right] \bigg \vert \Sigma_{t_{k-1}} \right] \nonumber\\
     &= \E\left[ \E_G\left[\sum_{i=1}^{T-t_{k-1}} \mathbbm{1}_{\{\tau_k\geq i\}}\cdot \left\|X_{t_{k-1}+i}\right\|^2_{\Sigma_{t_{k-1}}^{-1}} \bigg \vert \Sigma_{t_{k-1}},\{X_{t'}\}_{t'=t_{k-1}+1}^{T} \right] \bigg \vert \Sigma_{t_{k-1}} \right] \nonumber\\
     &= \E\left[ \sum_{i=1}^{T-t_{k-1}} \left\|X_{t_{k-1}+i}\right\|^2_{\Sigma_{t_{k-1}}^{-1}}\E_G\left[ \mathbbm{1}_{\{\tau_k\geq i\}} \bigg \vert \Sigma_{t_{k-1}},\{X_{t'}\}_{t'=t_{k-1}+1}^{T} \right] \bigg \vert \Sigma_{t_{k-1}} \right] \label{Thm32_eq1},
\end{align}
where the first line is by the tower rule of conditional expectation.  Given $X_t$, we denote $q_t^*$ as the purchasing  probability of  focal product at optimal  pricing  under  the  true  parameter, i.e.,
$$q_t^*:=1-F_f\left(p_f^*\left(V^*_{t,f},r^*(V^*_{t,a})\right)-V^*_{t,f}\right),$$
where $V^*_{t,f}=X^\top_t\theta^*_f$ and $V^*_{t,a}=X^\top_t\theta^*_a$. And we know that $\E[q_t^*]=q^*$ where the expectation is taken with respect to $X_t.$

From Lemma~\ref{price_1}, under event $\mathcal{E}_f\cap \mathcal{E}_a$, we know the focal product's purchase probability with LCB pricing is at least  $q_t^*$ almost surely. Notice that the probability $\mathbb{P}\left(\tau_k\geq i ,\mathcal{E}_f\cap\mathcal{E}_a\big \vert \{X_{t'}\}_{t'=t_{k-1}+1}^{T}\right)\leq \prod_{j=1}^{i-1}(1-q^*_{t_{k-1}+j})$ for $2\leq i\leq T-t_{k-1}$ and $\mathbb{P}\left(\tau_k\geq 1,\mathcal{E}_f\cap\mathcal{E}_a \big \vert \{X_{t'}\}_{t'=t_{k-1}+1}^{T}\right)=1$ by Lemma~\ref{price_1}, we have:

\begin{align}
    \eqref{Thm32_eq1}\leq& \E\left[\left\|X_{t_{k-1}+1}\right\|^2_{\Sigma_{t_{k-1}}^{-1}}+\sum_{i=2}^{T-t_{k-1}}\left(\left\|X_{t_{k-1}+i}\right\|^2_{\Sigma_{t_{k-1}}^{-1}}\cdot\prod_{j=1}^{i-1}(1-q^*_{t_{k-1}+j})\right) \bigg \vert \Sigma_{t_{k-1}} \right] \nonumber\\
    =& \E\left[\left\|X_{t_{k-1}+1}\right\|^2_{\Sigma_{t_{k-1}}^{-1}}\bigg \vert \Sigma_{t_{k-1}} \right]
    +\E\left[\sum_{i=2}^{T-t_{k-1}}\left(\left\|X_{t_{k-1}+i}\right\|^2_{\Sigma_{t_{k-1}}^{-1}}\cdot\E\left[\prod_{j=1}^{i-1}(1-q^*_{t_{k-1}+j})\big \vert X_{t_{k-1}+i},\Sigma_{t_{k-1}}\right] \right)\bigg \vert \Sigma_{t_{k-1}} \right] \nonumber\\
    =& \E\left[\left\|X_{t_{k-1}+1}\right\|^2_{\Sigma_{t_{k-1}}^{-1}}\bigg \vert \Sigma_{t_{k-1}} \right]
    +\E\left[\sum_{i=2}^{T-t_{k-1}}\left(\left\|X_{t_{k-1}+i}\right\|^2_{\Sigma_{t_{k-1}}^{-1}}\cdot\E\left[\prod_{j=1}^{i-1}(1-q^*_{t_{k-1}+j})\right] \right)\bigg \vert \Sigma_{t_{k-1}} \right] \nonumber\\
    =&\E\left[\left\|X_{t_{k-1}+1}\right\|^2_{\Sigma_{t_{k-1}}^{-1}}\bigg \vert \Sigma_{t_{k-1}} \right]
    +\E\left[\sum_{i=2}^{T-t_{k-1}}\left(\left\|X_{t_{k-1}+i}\right\|^2_{\Sigma_{t_{k-1}}^{-1}}\cdot(1-q^*)^{i-1}\right)\bigg \vert \Sigma_{t_{k-1}} \right] \nonumber\\
    =&\E\left[\left\|X_{t_{k-1}+1}\right\|^2_{\Sigma_{t_{k-1}}^{-1}}\bigg \vert \Sigma_{t_{k-1}} \right]
    +\E\left[\sum_{i=2}^{T-t_{k-1}}\left(\left\|X_{t_{k-1}+1}\right\|^2_{\Sigma_{t_{k-1}}^{-1}}\cdot(1-q^*)^{i-1}\right)\bigg \vert \Sigma_{t_{k-1}} \right] \nonumber\\
    =&\frac{1}{q^*}\E\left[\left\|X_{t_{k-1}+1}\right\|^2_{\Sigma_{t_{k-1}}^{-1}}\bigg \vert \Sigma_{t_{k-1}} \right]
\end{align}
where the the third line is by the independence between $1-q^*_{t_{k-1+j}}$ and $X_{t_{k-1+i}}$, $\Sigma_{t_{k-1}}$ for all $j=1,...,i-1$, the fourth line is by the i.i.d property for $q^*_{t_{k-1}+j}$ by Assumption~\ref{assp_X}, and the fifth line is again by the i.i.d property for $X_{t_{k-1+i}}$ and the its independence with $\Sigma_{t_{k-1}}^{-1}$. In fact the above analysis is very similar to the derivation of Wald's equation.

Thus, for (\ref{reg_sum}), it can be further expressed by
\begin{align}
    &144\eta \bar{\beta}^2\E_G\left[\sum_{k=1}^{N(T)}\sum_{t=t_{k-1}+1}^{t_{k}} \left\|X_t\right\|^2_{\Sigma_{t_{k-1}}^{-1}}\right]+144\eta \bar{\beta}^2\E_G\left[\sum_{t=t_{N(T)}+1}^{T} \left\|X_t\right\|^2_{\Sigma_{t_{N(T)}}^{-1}}\right] \nonumber\\
    &\leq \frac{144\eta \bar{\beta}^2}{q^*}\E\left[\sum_{k=1}^{N(T)} \left\|X_{t_{k-1}+1}\right\|^2_{\Sigma_{t_{k-1}}^{-1}}\right]+144\eta \bar{\beta}\E_G\left[\sum_{t=t_{N(T)}+1}^{t_{N(T)+1}} \left\|X_t\right\|^2_{\Sigma_{t_{N(T)}}^{-1}}\right] \nonumber\\
    &\leq \frac{144\eta \bar{\beta}^2}{q^*}\E\left[\sum_{k=1}^{N(T)+1} \left\|X_{t_{k-1}+1}\right\|^2_{\Sigma_{t_{k-1}}^{-1}}\right] \label{final_step_ub1},
\end{align}
where in the first inequality we suppose the selling periods can be extended from $T$ to $t_{N(T)+1}$. Thus, with $N(T)+1\leq T+1$, we can apply the elliptical potential Lemma~\ref{EPL_lemma}:
$$\sum_{k=1}^{N(T)+1} \left\|X_{t_{k-1}+1}\right\|^2_{\Sigma_{t_{k-1}}^{-1}}\leq 2d\log\left(\frac{d\lambda+(T+1)}{d\lambda}\right).$$
Plugging the above term in (\ref{final_step_ub1}) and setting $\lambda=1$, we complete the proof.
\qedwhite

\section{Proof of Section~\ref{sec_general}}

\renewcommand{\thesubsection}{C\arabic{subsection}}

To better present the proofs, we define the LCB/UCB estimators as follows
$$\bar{\theta}_{t-1,f}:=\argmax_{\theta\in\Theta_{t-1,f}} x_t^\top \theta,\quad \underline{\theta}_{t-1,f}:=\argmin_{\theta\in\Theta_{t-1,f}} x_t^\top \theta, $$
$$\bar{\theta}_{t-1,b}:=\argmax_{\theta\in\Theta_{t-1,b}} x_t^\top \theta,\quad \underline{\theta}_{t-1,b}:=\argmin_{\theta\in\Theta_{t-1,b}} x_t^\top \theta. $$

Further, we denote $v^*_{t,i}:=x_t^\top \theta^*_i$ for $i\in \{f,a,b\}$.

\paragraph{Proof of Lemma~\ref{reg_general_single}}

\proof{}\noindent
For simplicity, we only analyze the scenario when the bundling strategy is better, and the other scenario follows the same argument. When $A_t=u\neq A_t^*=b,$ we have
\begin{align*}
     r^*_b(v^*_{t,b})-r^*_u(v^*_{t,f},v^*_{t,a})
    & =\bar{r}^*_{t,u}-r^*_u(v^*_{t,f},v^*_{t,a})+r^*_b(v^*_{t,b})-\underline{r}^*_{t,b}+\underline{r}^*_{t,b}-\bar{r}^*_{t,u}\\
    & \leq \bar{r}^*_{t,u}-r^*_u(v^*_{t,f},v^*_{t,a})+r^*_b(v^*_{t,b})-\underline{r}^*_{t,b}.
\end{align*}
The last inequality holds because when \eqref{IDS_exploration} is adopted, we have $\underline{r}^*_{t,b}-\bar{r}^*_{t,u}\leq 0$.

For the first term, we have
\begin{align*}
    \bar{r}^*_{t,u}-r^*_u(v^*_{t,f},v^*_{t,a})
    & \leq |v^*_{t,f}-\bar{v}_{t,f}|+|v^*_{t,a}-\bar{v}'_{t,a}| \quad (\text{Lemma~\ref{Revenue_con}})\\
    & \leq |v^*_{t,f}-\bar{v}_{t,f}|+|v^*_{t,b}-\bar{v}_{t,b}|+|v^*_{t,f}-\underline{v}_{t,f}| \quad \text{(by definition of $\bar{v}'_{t,a}$)}\\
    &\leq \|x_t\|_{\Sigma_{t-1,f}^{-1}}\left(2\left\|\hat{\theta}_{t,f}-\theta^*_f\right\|_{\Sigma_{t-1,f}^{-1}}+\left\|\hat{\theta}_{t,f}-\bar{\theta}_{t,f}\right\|_{\Sigma_{t-1,f}^{-1}}+\left\|\hat{\theta}_{t,f}-\underline{\theta}_{t,f}\right\|_{\Sigma_{t-1,f}^{-1}}\right)\\
    & \ \ +\|x_t\|_{\Sigma_{t-1,b}^{-1}}\left(\left\|\hat{\theta}_{t,b}-\theta^*_b\right\|_{\Sigma_{t-1,b}^{-1}}+\left\|\hat{\theta}_{t,b}-\bar{\theta}_{t,b}\right\|_{\Sigma_{t-1,b}^{-1}}\right)\\
    & \leq \|x_t\|_{\Sigma_{t-1,f}^{-1}}\left(2\left\|\hat{\theta}_{t,f}-\theta^*_f\right\|_{\Sigma_{t-1,f}^{-1}}+2\beta\left(\Sigma_{t-1,f}^{-1}\right)\right) \\
   & \ \  +|x_t\|_{\Sigma_{t-1,b}^{-1}}\left(\left\|\hat{\theta}_{t,b}-\theta^*_b\right\|_{\Sigma_{t-1,b}^{-1}}+\beta\left(\Sigma_{t-1,b}^{-1}\right)\right) \quad \text{(by definition of the confidence set)}\\
    & \leq  4\|x_t\|_{\Sigma_{t-1,f}^{-1}}\beta\left(\Sigma_{t-1,f}^{-1}\right)+2\|x_t\|_{\Sigma_{t-1,b}^{-1}}\beta\left(\Sigma_{t-1,b}^{-1}\right)\quad \text{ (by $\mathcal{E}_{f}\cap \mathcal{E}_{b}$)}\\
    & \leq 4\bar{\beta}\left(\|x_t\|_{\Sigma_{t-1,f}^{-1}}+\|x_t\|_{\Sigma_{t-1,b}^{-1}}\right).
\end{align*}

For the second term, we have (in a similar manner as the first term)
\begin{align*}
    r^*_b(v^*_{t,b})-\underline{r}^*_{t,b} &\leq \left|v_{t,b}^*-\underline{v}_{t,b}\right| \quad \text{(Lemma~\ref{Rev_11})}\\
    & \leq 2\bar{\beta}\|x_t\|_{\Sigma_{t-1,b}^{-1}}.
\end{align*}

Thus, the proof is completed by combining above two inequalities
$$r^*_b(v^*_{t,b})-r^*_u(v^*_{t,f},v^*_{t,a})\leq  6\bar{\beta}\left(\|x_t\|_{\Sigma_{t-1,f}^{-1}}+\|x_t\|_{\Sigma_{t-1,b}^{-1}}\right). $$
\qedwhite

\paragraph{Proof of Lemma~\ref{reg_gen_choice_sum}}

\proof{}\noindent
We need to first introduce some new notations. Denote
$$\Sigma_{t,f}'=\sum_{t'\in \mathcal{T}'_{t,f}}x_{t'}x_{t'}^\top+\lambda I,$$
$$\Sigma_{t,b}'=\sum_{t'\in \mathcal{T}'_{t,b}}x_{t'}x_{t'}^\top+\lambda I,$$
where $\mathcal{T}'_{t,f}$ and $\mathcal{T}'_{t,b}$ are the set of \textbf{exploration} periods gathering corresponding samples before $t$ when (5) is adopted (recall $\mathcal{T}_{t,f}$ and $\mathcal{T}_{t,b}$ are the set of all periods gathering corresponding samples before $t$). Then a simple implication is that both
$\Sigma_{t,f} - \Sigma_{t,f}'$ and  $ \Sigma_{t,b}-\Sigma_{t,b}'$ are positive semi-definite matrices.

For brevity, we use
$\E_{G'}[\cdot]:=\E\left[(\cdot)\mathbbm{1}_{\mathcal{E}_f\cap \mathcal{E}_b}\right]$, then
\begin{align}
\E\left[\sum_{t=1}^T \left\vert r^*_b\left(v^*_{t,b}\right)-r^*_u\left(v^*_{t,f},v^*_{t,a}\right)\right\vert \cdot \mathbbm{1}_{A_t\neq A_t^*}\right]     &\leq6\bar{\beta}\E_{G'}\left[\sum_{t\in \mathcal{T}'_{T,f}\cup \mathcal{T}'_{T,b}} \left(\|x_t\|_{\Sigma_{t-1,f}^{-1}}+\|x_t\|_{\Sigma_{t-1,b}^{-1}}\right)\right] \nonumber \\
    &\leq6\bar{\beta}\E\left[\sum_{t\in \mathcal{T}'_{T,f}\cup \mathcal{T}'_{T,b}} \left(\|x_t\|_{\Sigma_{t-1,f}^{-1}}+\|x_t\|_{\Sigma_{t-1,b}^{-1}}\right)\right] \nonumber\\
    &\leq 6\sqrt{2}\bar{\beta}\sqrt{T}\sqrt{\E\left[\sum_{t\in \mathcal{T}'_{T,f}\cup \mathcal{T}'_{T,b}} \left(\|x_t\|^2_{(\Sigma'_{t-1,f})^{-1}}+\|x_t\|^2_{(\Sigma'_{t-1,b})^{-1}}\right)\right]}.\label{reg_choice_sum}
\end{align}
Here the first line comes from Lemma~\ref{reg_general_single}, and the last line is by Cauchy-Schwartz inequality and the fact that $|\mathcal{T}'_{T,f}\cup \mathcal{T}'_{T,b}|\leq T$.

By decision rule \eqref{IDS_exploration}, we know
$$\|x_t\|^2_{(\Sigma'_{t-1,f})^{-1}}\geq \|x_t\|^2_{(\Sigma'_{t-1,b})^{-1}}, \quad \forall t\in \mathcal{T}'_{t,f},$$
$$\|x_t\|^2_{(\Sigma'_{t-1,f})^{-1}}< \|x_t\|^2_{(\Sigma'_{t-1,b})^{-1}}, \quad \forall t\in \mathcal{T}'_{T,b}.$$
Thus,
\begin{align*}
    \E\left[\sum_{t\in \mathcal{T}'_{T,f}\cup \mathcal{T}'_{T,b}} \left(\|x_t\|^2_{(\Sigma'_{t-1,f})^{-1}}+\|x_t\|^2_{(\Sigma'_{t-1,b})^{-1}}\right)\right]
    & \leq \E\left[2\sum_{t\in \mathcal{T}'_{T,f}}\|x_t\|^2_{(\Sigma'_{t-1,f})^{-1}}+2\sum_{t\in \mathcal{T}'_{T,b}}\|x_t\|^2_{(\Sigma'_{t-1,b})^{-1}}\right]\\
&    \leq 8d\log\left(\frac{d\lambda+T}{d\lambda}\right),
\end{align*}
where the last inequality is by Lemma~\ref{EPL_lemma} and the fact that $|\mathcal{T}'_{T,f}|, |\mathcal{T}'_{T,b}|\leq T$. By plugging the last line above into (\ref{reg_choice_sum}), we complete the proof.
\qedwhite

\paragraph{Proof of Lemma~\ref{Unbund_price_reg}}
\proof{}\noindent
Conditional on a fixed set of unbundling periods $\mathcal{T}_{T,f}$, with similar argument as in the proof of Theorem~\ref{UB_purepricing_2}, the expected pricing regret can be bounded by
\begin{align}
    \sum_{t\in \mathcal{T}_{T,f}}\E\left[q_{t,f}\left(6\bar{\beta}\left\|x_t\right\|_{\Sigma_{t-1,a}^{-1}}+\eta\bar{\beta}^2\left\|x_t\right\|_{\Sigma_{t-1,a}^{-1}}^2\right)\right]
   = \E\left[\sum_{t\in \mathcal{T}_{T,f}} \mathbbm{1}_{\{d_{t,f}=1\}}\left(6\bar{\beta}\left\|x_t\right\|_{\Sigma_{t-1,a}^{-1}}+\eta\bar{\beta}^2\left\|x_t\right\|_{\Sigma_{t-1,a}^{-1}}^2\right)\right] \label{GE_reg_sum_new_3},
\end{align}

By noting that $\sum_{t\in \mathcal{T}_{T,f}} \mathbbm{1}_{\{d_{t,f}=1\}}(\cdot)=\sum_{t\in \mathcal{T}_{T,a}}(\cdot)$, i.e., we only sum over the periods with focal purchase, $(\ref{GE_reg_sum_new_3})$ can be further expressed as:
\begin{align*}
    (\ref{GE_reg_sum_new_3})=&\E\left[\sum_{t\in \mathcal{T}_{T,a}}\left(6\bar{\beta}\left\|x_t\right\|_{\Sigma_{t-1,a}^{-1}}+\eta\bar{\beta}^2\left\|x_t\right\|_{\Sigma_{t-1,a}^{-1}}^2\right)\right]\\
    \leq & \E\left[\left(6\bar{\beta}\sqrt{|\mathcal{T}_{T,a}|\sum_{t \in \mathcal{T}_{T,a}}\left\|x_{t}\right\|_{\Sigma_{t-1,a}^{-1}}^2}+\eta\bar{\beta}^2\sum_{t\in \mathcal{T}_{T,a}}\left\|x_{t}\right\|_{\Sigma_{t-1,a}^{-1}}^2\right)\right],
\end{align*}
for any $\mathcal{T}_{T,f}$. Thus, with $|\mathcal{T}_{T,a}|\leq T$ we can finally apply the elliptical potential lemma again and take the expectation over $\mathcal{T}_{T,f}$ to complete the proof.
\qedwhite

\paragraph{Proof of Lemma~\ref{Reg_price_bundle}}
\proof{}\noindent
We first check the single period pricing regret under $\mathcal{E}_b$. Given $\Sigma_{t-1,b}$ and $x_t$, the used pricing policy will cause:
\begin{align*}
    &r_b^*(v^*_{t,b})-r_b^*\left(p_b^*(x_t^\top \hat{\theta}_{t,b});v^*_{t,b}\right)\\
    &\leq \eta\left|x_t^\top \hat{\theta}_{t,b}-v^*_{t,b}\right|^2 \quad \text{ (Lemma~\ref{Rev_11}) }\\
    &\leq \eta\bar{\beta}^2\left\|x_t\right\|^2_{\Sigma_{t-1,b}^{-1}}.
\end{align*}

And the total expected regret is:
\begin{align*}
    &\E\left[ \sum_{t\in \mathcal{T}_{T,b}} \eta\bar{\beta}^2\left\|x_t\right\|^2_{\Sigma_{t-1,b}^{-1}} \right]\\
    &\leq 2d\eta \bar{\beta}^2\log\left(\frac{d\lambda+T}{d\lambda}\right),
\end{align*}
by using the elliptical potential lemma~\ref{EPL_lemma} and the fact $|\mathcal{T}_{T,b}|\leq T$.
\qedwhite

\section{Proof of Section~\ref{sec_one_switch}}

\label{sec_proof_one_switch}

\renewcommand{\thesubsection}{D\arabic{subsection}}

\paragraph{UCB and LCB of revenues.}

\begin{lemma}
\label{lemma:one_switch_good_event}
Under Assumptions ~\ref{assp_add},~\ref{assp_bound},~\ref{assp_distr},~\ref{assp_X}, at period $t=1,...,T$,
$$\mathbb{P}\left(\E_X[r^*_u(X^\top\theta^*_f,X^\top \theta^*_a)]\notin [\underline{r}^*_{t,u},\bar{r}^*_{t,u}]\right)\leq \frac{4}{T^2}$$
$$\mathbb{P}\left(\E_X[r^*_b(X^\top\theta^*_b)]\notin [\underline{r}^*_{t,b},\bar{r}^*_{t,b}]\right)\leq \frac{4}{T^2}$$
\end{lemma}
\proof{}\noindent
Notice that the sequence
$$\left\{\sum_{t'=1}^t \left(r^*_u(X_{t'}^\top\theta_f^*,X_{t'}^\top\theta_a^*)-\mathbb{E}_X[r_u^*(X^\top\theta_f^*,X^\top\theta_a^*)]\right)\right\}_{t=1,..,T}$$
is a martingale with bounded difference
$$\left| r^*_u(X_{t'}^\top\theta_f^*,X_{t'}^\top\theta_a^*)-\mathbb{E}_X[r_u^*(X^\top\theta_f^*,X^\top\theta_a^*)]\right|\leq 2\Bar{p}.$$

By Azuma inequality, for any $t=1...,T$
\begin{equation}
\label{one_switch_prob_1}
   \mathbb{P}\left( \left|\frac{1}{t} \sum_{t'=1}^t r^*_u(X_{t'}^\top\theta_f^*,X_{t'}^\top\theta_a^*)-\mathbb{E}_X[r_u^*(X^\top\theta_f^*,X^\top\theta_a^*)]\right| \geq 4\Bar{p}\sqrt{\frac{\log T}{t}}\right)\leq \frac{2}{T^2},
\end{equation}

Further, by Lemma~\ref{Rev_2} and Lemma~\ref{Revenue_con}, for any  $x\in \mathcal{X}$,
\begin{align*}
  &|r^*_u(x^\top \hat{\theta}_{t+1,f}, x^\top \hat{\theta}_{t+1,a})-r^*_u(x^\top \theta_f^*, x^\top \theta_a^*)|\\
  \leq &|x^\top(\hat{\theta}_{t+1,f}-\theta_f^*)|+|x^\top(\hat{\theta}_{t+1,a}-\theta_a^*)|\\
  \leq &\|x\|_{\Sigma_{t,f}^{-1}} \|\hat{\theta}_{t+1,f}-\theta_f^*\|_{\Sigma_{t,f}}+\|x\|_{\Sigma_{t,a}^{-1}} \|\hat{\theta}_{t+1,a}-\theta_a^*\|_{\Sigma_{t,a}}.
\end{align*}
Replace $x=X_{t'}$, $t'=1,..,t$, we have
\begin{align*}
    &\frac{1}{t}\sum_{t'=1}^{t}\left| r^*_u(X_{t'}^\top \hat{\theta}_{t+1,f}, X_{t'}^\top \hat{\theta}_{t+1,a})-r^*_u(X_{t'}^\top \theta_f^*, X_{t'}^\top \theta_a^*)\right|\\
    \leq & \frac{1}{t}\sum_{t'=1}^{t} \left(\|X_{t'}\|_{\Sigma_{t,f}^{-1}} \|\hat{\theta}_{t+1,f}-\theta_f^*\|_{\Sigma_{t,f}}+\|X_{t'}\|_{\Sigma_{t,a}^{-1}} \|\hat{\theta}_{t+1,a}-\theta_a^*\|_{\Sigma_{t,a}}\right)\\
    \leq & \frac{1}{t}\sum_{t'=1}^{t} \|X_{t'}\|_{\Sigma_{t'-1,a}^{-1}} \left( \|\hat{\theta}_{t+1,f}-\theta_f^*\|_{\Sigma_{t,f}}+ \|\hat{\theta}_{t+1,a}-\theta_a^*\|_{\Sigma_{t,a}}\right)
\end{align*}
where the last line is because $\Sigma_{t'-1,a}^{-1}\geq \Sigma_{t,a}^{-1}\geq \Sigma_{t,f}^{-1}$ for all $t'\leq t$. Thus,
\begin{align*}
&\mathbb{P}\left(\E_X[r^*_u(X^\top\theta^*_f,X^\top \theta^*_a)]\notin [\underline{r}^*_{t,u},\bar{r}^*_{t,u}]\right)\\
&\leq \mathbb{P}\left(\frac{1}{t}\left|    t\mathbb{E}_X[r_u^*(X^\top\theta_f^*,X^\top\theta_a^*)]- \sum_{t'=1}^{t} r^*_u(X_{t'}^\top \hat{\theta}_{t,f}, X_{t'}^\top \hat{\theta}_{t,a}) \right|\geq  \frac{2\Bar{\beta}}{t}\sum_{t'=1}^{t} \|X_{t'}\|_{\Sigma_{t'-1,a}^{-1}}+4\Bar{p}\sqrt{\frac{\log T}{t}}    \right)\\
&\leq \mathbb{P}\left(\frac{1}{t}\sum_{t'=1}^{t}\left| r^*_u(X_{t'}^\top \hat{\theta}_{t+1,f}, X_{t'}^\top \hat{\theta}_{t+1,a})-r^*_u(X_{t'}^\top \theta_f^*, X_{t'}^\top \theta_a^*)\right|\geq \frac{2\Bar{\beta}}{t}\sum_{t'=1}^{t} \|X_{t'}\|_{\Sigma_{t'-1,a}^{-1}}  \right)\\
&+\mathbb{P}\left(\left| \frac{1}{t}\sum_{t'=1}^t r^*_u(X_{t'}^\top\theta_f^*,X_{t'}^\top \theta_a^*)-\mathbb{E}_X[r_u^*(X^\top\theta_f^*,X^\top\theta_a^*)]\right| \geq 4\Bar{p}\sqrt{\frac{\log T}{t}} \right)\\
 &\leq  \mathbb{P}\left(\|\hat{\theta}_{t+1,f}-\theta_f^*\|_{\Sigma_{t,f}}\geq \Bar{\beta} \right) +\mathbb{P}\left(\|\hat{\theta}_{t+1,a}-\theta_a^*\|_{\Sigma_{t,a}}\geq \Bar{\beta} \right)\\
 &+\mathbb{P}\left(\left| \frac{1}{t}\sum_{t'=1}^t r^*_u(X_{t'}^\top\theta_f^*,X_{t'}^\top \theta_a^*)-\mathbb{E}_X[r_u^*(X^\top\theta_f^*,X^\top\theta_a^*)]\right| \geq 4\Bar{p}\sqrt{\frac{\log T}{t}} \right)\\
 &\leq \frac{4}{T^2}
\end{align*}
where the last line is by Corollary~\ref{MLEbound} and Corollary~\ref{ub_beta} with \eqref{one_switch_prob_1}. And we can get the same bound for bundling through above arguments.
\qedwhite

\subsection{Regret decomposition: strategy regret, pricing regret and good event}
Like the general case, we can decompose the regret into strategy regret and pricing regret. The only difference is that the strategy regret now is the the gap of \textbf{expected} revenues. Without loss of generality, we assume the optimal selling strategy is bundling, then when choosing unbundling:

\begin{multline*}
    \E_X\left[r^*_b\left(X^\top \theta^*_b \right)-r_u\left(p_{t,f},p_{t,a};X^\top \theta^*_f,X^\top \theta^*_a\right)\right]\\=\underbrace{\E_X\left[r^*_b\left(X^\top \theta^*_b\right)-r^*_u\left(X^\top \theta^*_f,X^\top \theta^*_a\right)\right]}_{\text{Strategy regret}}+\underbrace{\E_X\left[r^*_u\left(X^\top \theta^*_f,X^\top \theta^*_a\right)-r_u\left(p_{t,f},p_{t,a};X^\top \theta^*_f,X^\top \theta^*_a\right)\right]}_{\text{Pricing regret}},
\end{multline*}
where $\E_X[\cdot]$ is the expectation over $X$.

 \paragraph{Regret from failure of good event .} First notice that the event $\mathcal{E}_f \cap \mathcal{E}_a\cap \mathcal{E}_b$ has probability at least $1-\frac{3}{T}$ by Lemma~\ref{lemma_high_prob_1}. And then define
 $$\mathcal{E}:=\left\{\E_X[r^*_u(X^\top\theta^*_f,X^\top \theta^*_a)\in [\underline{r}^*_{t,u},\bar{r}^*_{t,u}], \E_X[r^*_b(X^\top\theta^*_b)]\in [\underline{r}^*_{t,b},\bar{r}^*_{t,b}] \quad \text{for} \  t=1,...,T \right\}\cap \mathcal{E}_f \cap \mathcal{E}_a\cap \mathcal{E}_b,$$
 by Lemma~\ref{lemma:one_switch_good_event} with union bound, we have
$$\mathbb{P}\left(\mathcal{E}\right)\geq 1-\frac{11}{T}.$$

 By noticing the single regret is bounded by $2\bar{p}$, the total expected regret caused by failure of good event can be bounded by $22\bar{p}$.

\subsection{Strategy regret under good event}
\paragraph{Exploitation periods.} Under good event $\mathcal{E}$, following the same argument in general case, the true optimal expected revenues will be included in the LCB and UCB and there will be no regret caused in exploitation periods under the good event.

\paragraph{Exploration periods} We first check the single period regret for one sample path:
\begin{lemma}
\label{lemma:single_period_sto_single_reg}
With $\mathcal{E}$, if the algorithm chooses exploration (thus keeps using unbundling), the single period strategy regret at $t+1$ can be bounded by
$$16\Bar{p}\sqrt{\frac{\log T}{t}}+\frac{8\Bar{\beta}}{t}\sum_{t'=1}^t \|X_{t'}\|_{\Sigma^{-1}_{t'-1,a}}.$$
\end{lemma}
We omit the proof here since it is similar as Lemma~\ref{reg_general_single} with Lemma~\ref{lemma:one_switch_good_event} except that we directly estimating $\theta_a$ instead of using $\theta_b-\theta_f$. Then after checking the expectation $\E\left[ \sum_{t'=1}^t \|X_{t'}\|_{\Sigma^{-1}_{t'-1,a}}\mathbbm{1}_{\mathcal{E}}\right]$, we can bound the single period expected regret.

\begin{lemma}
\label{lemma:single_period_exp_single_reg}
For $t=1,...,T-1$, with $\lambda\geq 1$
$$\E\left[\sum_{t'=1}^t \|X_{t'}\|_{\Sigma^{-1}_{t'-1,a}}\mathbbm{1}_{\mathcal{E}}\right]\leq \sqrt{\frac{2dt}{q^*} \log\left(\frac{d\lambda+T}{d\lambda}\right)}$$
\end{lemma}
\proof{}\noindent
By Holder's inequality,
\begin{align}
    &\E\left[\sum_{t'=1}^t  \|X_{t'}\|_{\Sigma^{-1}_{t'-1,a}}\mathbbm{1}_{\mathcal{E}}\right] \nonumber\\
    &\leq \E\left[\sum_{t'=1}^t  \|X_{t'}\|_{\Sigma^{-1}_{t'-1,a}}\right] \nonumber\\
    &\leq  \sqrt{t
   E\left[ \sum_{t'=1}^t \|X_{t'}\|^2_{\Sigma^{-1}_{t'-1,a}}\right]} \label{eqn:one_switch_single_bound}
\end{align}
Then by a similar argument in Theorem~\ref{UB_purepricing_1}, we define $\mathcal{T}'_{t,a}$ as the time index set when a focal purchase takes place at time $t$, we have
\begin{align*}
    &\mathbb{E}\left[\sum_{t'=1}^t\|X_{t'}\|^2_{\Sigma_{t'-1,a}^{-1}}\right]\\
    &\leq \frac{1}{q^*}\mathbb{E}\left[\sum_{t'\in \mathcal{T}'_{t,a}} \|X_{t'}\|^2_{\Sigma_{t'-1,a}^{-1}}\right]\\
    &\leq \frac{2d}{q^*}\log\left(\frac{d\lambda+T}{d\lambda}\right),
\end{align*}
where the last inequality is by elliptical potential lemma~\ref{EPL_lemma}. By plugging above inequality in \eqref{eqn:one_switch_single_bound}, we complete our proof.

\qedwhite

Thus by combining Lemma~\ref{lemma:single_period_sto_single_reg} and   Lemma~\ref{lemma:single_period_exp_single_reg} the total expected regret bound caused in exploration periods can be bounded by:
\begin{lemma}
\label{One-switching_choice_ub}
With $\mathcal{E}$, the total expected strategy regret caused in exploration periods can be upper bounded by for any $\lambda\geq 1$,

$$16\Bar{p}\sqrt{T\log T}+8\Bar{p}\sqrt{\frac{2dT}{q^*}\log\left(\frac{d\lambda+T}{d\lambda}\right) }$$
\end{lemma}
\proof{}\noindent
By combining Lemma~\ref{lemma:single_period_sto_single_reg} and   Lemma~\ref{lemma:single_period_exp_single_reg}, the single period expected regret at time $t+1$ can be bounded by
$$16\Bar{p}\sqrt{\frac{\log T}{t}}+8\Bar{p}\sqrt{\frac{2d}{q^*t}\log\left(\frac{d\lambda+T}{d\lambda}\right) }.$$
Noting $\sum_{t=1}^T 1/\sqrt{t}\leq \sqrt{T+1}$ and the total number of explorations is bounded by $T$, by summing above for $t=1,..,T$, and we complete the proof.
\qedwhite

%%%%%%%%%%%%%%%%%%%%%%%%%%%%%%%%%%%%%%%%%%%%%%%%%%%%%%%%%%%%%%%%%%%%%%

% Samples of sectioning (and labeling) in MOOR.
% NOTE: (1) all section levels end with a period,
%       (2) capitalization is as shown (sentence style, not title style).
%
%\section{Introduction.}\label{intro} %%1.
%\subsection{Duality and the classical EOQ problem.}\label{class-EOQ} %% 1.1.
%\subsection{Outline.}\label{outline1} %% 1.2.
%\subsubsection{Cyclic schedules for the general deterministic SMDP.}
%  \label{cyclic-schedules} %% 1.2.1
%\section{Problem description.}\label{problemdescription} %% 2.

% Text of your paper here

% Appendix here
% Options are (1) APPENDIX (with or without general title) or
%             (2) APPENDICES (if it has more than one unrelated sections)
% Outcomment the appropriate case if necessary
%
% \begin{APPENDIX}{<Title of the Appendix>}
% \end{APPENDIX}
%
%   or
%
% \begin{APPENDICES}
% \section{<Title of Section A>}
% \section{<Title of Section B>}
% etc

% Acknowledgments here

% Enter the text of acknowledgments here

% References here (outcomment the appropriate case)

% CASE 1: BiBTeX used to constantly update the references
%   (while the paper is being written).
%\bibliographystyle{informs2014} % outcomment this and next line in Case 1
%\bibliography{<your bib file(s)>} % if more than one, comma separated

% CASE 2: BiBTeX used to generate mypaper.bbl (to be further fine tuned)
%\input{mypaper.bbl} % outcomment this line in Case 2

\end{document}